\documentclass[3p,times]{elsarticle}

\usepackage{ecrc}
\usepackage{enumitem}
\usepackage{amsmath}
\usepackage{multirow}
\usepackage{multicol}
\usepackage{color}
\usepackage{subfig} 
\usepackage{booktabs}

\volume{00}

\firstpage{1}

\journalname{Information Fusion}

\runauth{Silu He, Peng Shen, Pingzhen Xu, Qinyao Luo, Haifeng Li}

\jid{procs}

\jnltitlelogo{Information Fusion}

\CopyrightLine{2011}{Published by Elsevier Ltd.}

\usepackage{amssymb}

\usepackage[figuresright]{rotating}

\begin{document}

\begin{frontmatter}



\dochead{}

\title{STDCformer: A Transformer-Based Model with a Spatial-Temporal Causal De-Confounding Strategy for Crowd Flow Prediction}


\author{Silu He, Peng Shen, Pingzhen Xu, Qinyao Luo, Haifeng Li\corref{cor1}}

\address{School of Geosciences and Info-Physics, Central South University, Changsha, 410083, Hunan, China}

\cortext[cor1]{Corresponding author}

\begin{abstract}

Crowd Flow Prediction is critical to urban management, with the goal of capturing the arrival and departure characteristics of crowd movements under different spatial and temporal distributions, which is fundamentally a spatial-temporal prediction task. Existing works typically treat spatial-temporal prediction as the task of learning a function $F$  to transform historical observations to future observations. We further decompose this cross-time transformation into three processes: (1) Encoding ($E$): learning the intrinsic representation of observations, (2) Cross-Time Mapping ($M$): transforming past representations into future representations, and (3) Decoding ($D$): reconstructing future observations from the future representations. From this perspective, spatial-temporal prediction can be viewed as learning  $F = E \cdot M \cdot D$, which includes learning the space transformations $\left\{{E},{D}\right\}$ between the observation space and the hidden representation space, as well as the spatial-temporal mapping $M$ from future states to past states within the representation space. This leads to two key questions: \textbf{Q1: What kind of representation space allows for mapping the past to the future? Q2: How to achieve mapping the past to the future within the representation space?} To address Q1, we propose a Spatial-Temporal Backdoor Adjustment strategy, which learns a Spatial-Temporal De-Confounded (STDC) representation space and estimates the de-confounding causal effect of historical data on future data. This causal relationship we captured serves as the foundation for subsequent spatial-temporal mapping. To address Q2, we design a Spatial-Temporal Embedding (STE) that fuses the information of temporal and spatial confounders, capturing the intrinsic spatial-temporal characteristics of the representations. Additionally, we introduce a Cross-Time Attention mechanism, which queries the attention between the future and the past to guide spatial-temporal mapping. Finally, we integrate the process of learning the STDC representation space and the spatial-temporal mapping into an $E$-$M$-$D$ skeleton for spatial-temporal prediction. The skeleton is further instantiated with a Transformer model, building a Transformer model with Spatial-Temporal De-Confounding Strategy (STDCformer). Experiments on two real-world datasets demonstrate that STDCformer achieves state-of-the-art predictive performance and exhibits stronger out-of-distribution generalization capabilities.

\end{abstract}


\end{frontmatter}


\section{Introduction}
Crowd flow prediction aims to use historical flow data in a region to predict the inflow and outflow of during future time periods. It is a typical spatial-temporal prediction task that plays a crucial role in urban planning, traffic management, public safety, and other fields \cite{RN1, RN2}. The essence of spatial-temporal prediction lies in capturing the mapping $F_{past\rightarrow future}$ from historical data to future data, enabling the inference of future from historical observations. In recent years, a series of spatial-temporal prediction models have emerged with the goal of solving $F_{past\rightarrow future}$, and the deep learning-based models have become mainstream. Among these, models based on Spatial-Temporal Graph Neural Networks (STGNNs) and Spatial-Temporal Transformers (ST Transformers) have demonstrated exceptional performance in various spatial-temporal prediction tasks, such as traffic flow prediction and crowd flow prediction \cite{RN3, RN4}. These two types of models follow a unified paradigm, where temporal representation modules (e.g., RNN\cite{RN5,RN6,RN7,RN8,RN9,RN10,RN11}, CNN\cite{RN12,RN13,RN14,RN15,RN16}, and Transformer\cite{RN17,RN18,RN19,RN20}) and spatial representation modules (e.g., CNNs \cite{RN21}, GNNs \cite{RN5,RN6,RN7,RN10,RN12,RN13,RN15,RN15,RN16}, and Transformers \cite{RN8,RN17,RN20,RN22}) are used to separately capture the temporal and spatial representations from the observed data. A spatial-temporal fusion module then combines these representations to obtain the final representation, which is used to infer unobserved spatial-temporal data.

$F_{past\rightarrow future}$ needs to be realized through three subprocesses. First, the encoder $E$ is used to encode historical observations into representations in a latent space. Second, a cross-time mapping $M$ is applied to transform the representation of past states into the representation of future states. Third, the decoder $D$ projects the future representations back into the observation space to reconstruct the future observations. From this perspective, the objective of learning $F_{past\rightarrow future}$ is to learn $F = E \cdot M \cdot D$, which includes the space transformation ${E,D}$ between the observation space and the representation space, as well as the spatial-temporal mapping ${M}$ from future states to past states within the representation space. The purpose of space transformation is to identify a representation space that can capture the essential information in the observations, serving as the foundation for ensuring the feasibility of the subsequent spatial-temporal mapping. The purpose of spatial-temporal mapping is to learn the intrinsic, dynamic transformation relationships between the past and future from their essential representations. These two processes lead to two key questions behind spatial-temporal prediction: \textbf{Q1: What kind of representation space allows for mapping the past to the future? Q2: How to achieve mapping the past to the future within the representation space?} The answers to these questions and the corresponding methods are as follows.

\textbf{A1: The premise for mapping the past to the future is the accurate transmission of causal effects from the past to the future. Therefore, the mapping from the past to the future can be achieved in a Spatial-Temporal De-Confounded (STDC) representation space.}

The premise of inferring the future from the past is based on the assumption that there exists some form of "influence" of the past on the future, which determines the mapping relationship between the past and future. Historical data and future data are not simply related as input and output, but implicitly involve assumptions about the causal relationship between the past and future. Therefore, fusing the representation of causality can capture the essential information within the observational data. Under this assumption, such "influence" can be characterized through causal effects. Existing methods typically build neural network models to fit the associations $P(Future|Past)$ between historical and future data distributions from the observational data to learn the representation space. However, when uncontrolled confounding bias \cite{RN23} is present, the model fails to capture the true causal effect $P(Future|do(Past))$.

\begin{figure}[h]
\centering
\includegraphics[width=0.6\textwidth]{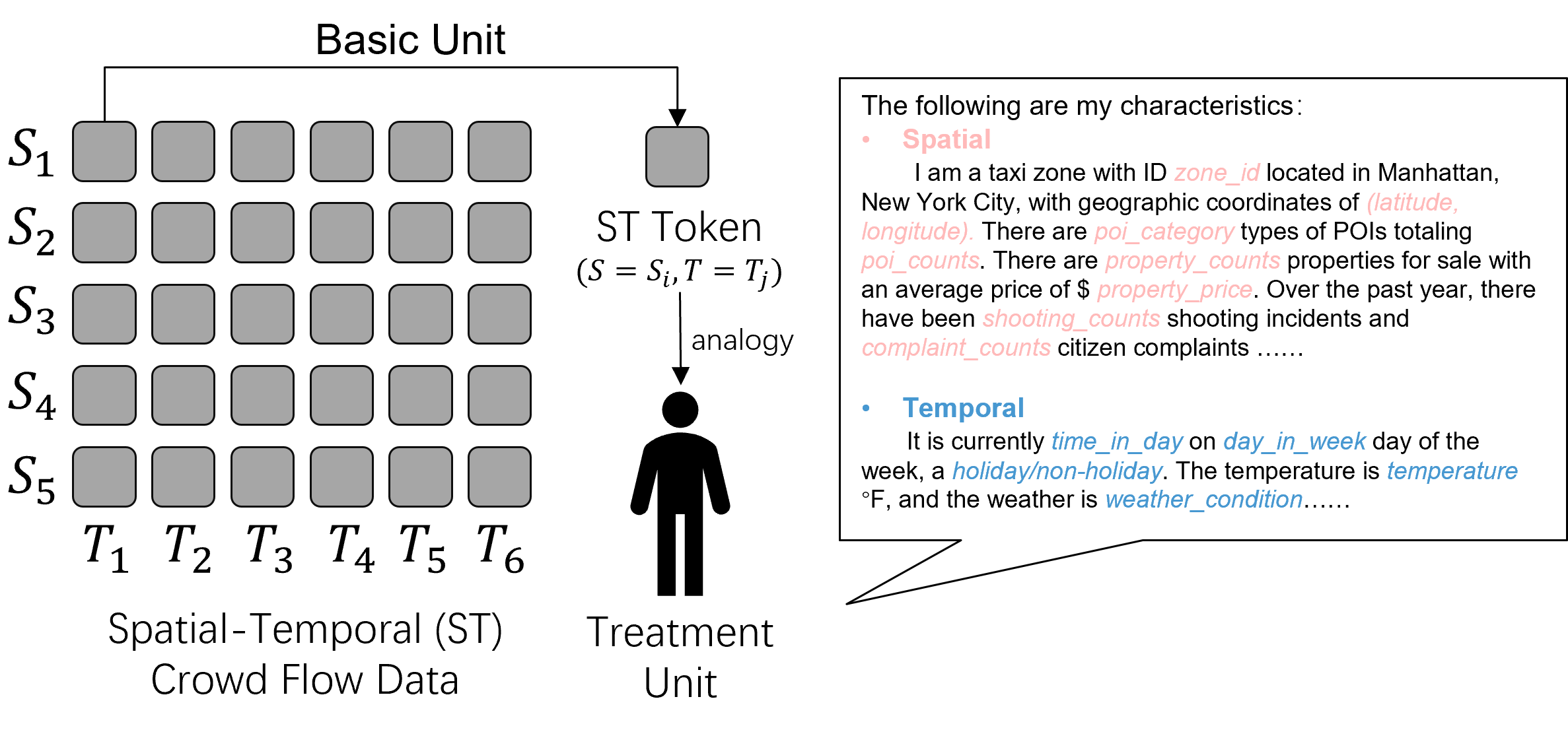}
\caption{The characteristics of Spatial-Temporal Tokens in crowd flow prediction task.}
\label{fig: fig1}
\end{figure}

As shown in Figure \ref{fig: fig1}, the basic unit in spatial-temporal data is analogous to a token in text, which serves as the trial unit for estimating $P(Future|do(Past))$, referred to as the ST token (STT), represented as $STT_{ij}: S = S_i, T = T_j$. It can be observed that, similar to different participants with varying physical conditions in a drug trial, each STT also has temporal and spatial characteristics. These characteristics reflect the attributes of the spatial region $S_i$ (e.g., functionality, travel cost, travel safety, etc.) and the temporal window $T_j$ (e.g., travel necessity, travel suitability, etc.), which are potential factors influencing human movement. These characteristics simultaneously affect both historical and future crowd flow observations, serving as confounders in the spatial-temporal prediction process. If the distribution bias of these confounding factors in the sample data is ignored, the model will learn incorrect patterns.

Estimating the correct causal effect requires eliminating the distributional bias of confounders, a process known as de-confounding. In this paper, we argue that the ideal spatial-temporal representation space should enable de-confounding, i.e., learn a Spatial-Temporal De-Confounded (STDC) representation space. The challenge of spatial-temporal observational data de-confounding lies in simulating the intervention operation $do(Past)$ on the observed samples, with the key being sample stratification and confounder control. Sample stratification entails classifying samples into treatment groups based on their differences, while confounder control then re-weight the samples in each stratum to estimate the true causal effect. However, due to the complexity of spatial-temporal activities and the limited nature of observational data, confounders are often unobservable, and the distribution of such hidden confounders in the observational samples is unknown, which makes it difficult to directly measure the differences between samples, consequently, challenging to accurately pre-define the number of sample strata and their weights. In order to obtain the representation of hidden confounders, existing methods often learn the representations confounders from the original observation time series \cite{RN24,RN25}. However, these abstract representations are hard to be aligned to the specific semantics in the real world, and the separability between different strata heavily depends on the sampling and quality of the observation data. Some methods also introduce Point-of-Interest (POI) data to learn the classification of confounding representations \cite{RN26}, but they need to pre-define the number of categories k of confounding and their distribution. The ideal hidden confounding stratification in crowd data should accurately indicate the key influence factors behind crowd movements, and different confounding representations should be naturally distinguishable and assigned appropriate weights. To approximate this ideal hidden confounder stratification and allow for subsequent control, this paper proposes a Spatial-Temporal Backdoor Adjustment strategy, specifically:

(1) To improve the reliability and semantics of confounders' representations, we collected auxiliary information that can separately characterize the temporal and spatial characteristics of STTs based on the analysis of human movement behavior. This auxiliary information is fused into the prediction model to learn the representations of hidden confounders. Compared to one-dimensional time series data from uncontrollable sampling process, this type of information has more stable quality and clearer semantics.

(2) To reduce the bias caused by assumptions on the distribution of confounding factors, and based on the two most critical attributes of human mobility, {{"When", "Where"}}, we naturally categorize the confounders into temporal and spatial confounders. This approach avoids the need to predefine the number of clusters k while ensuring the completeness of the stratification. Additionally, both temporal and spatial confounders are fed into learnable modules to derive the weights, eliminating the need to assume a predefined distribution for the confounders.

\textbf{A2: The intrinsic relationship between the past and the future depends on the characteristics of their spatial-temporal contexts, which can be portrayed by spatial and temporal confounders. Therefore, the representations of confounders can be used for querying the relationships between the past and future.}

\begin{figure}[h]
\centering
\includegraphics[width=0.6\textwidth]{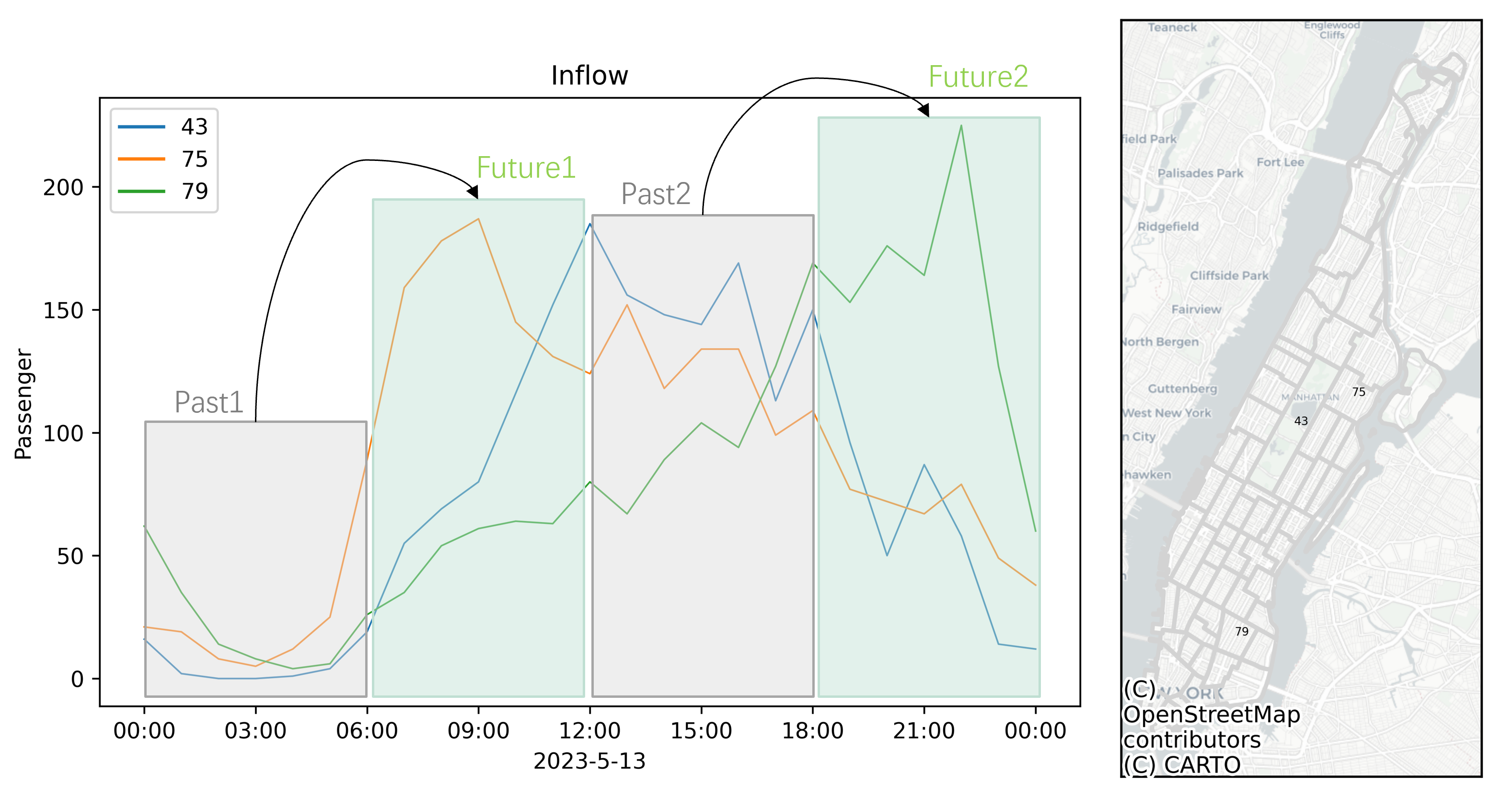}
\caption{Difference between historical and future flow under different time and region sampling.}
\label{fig: fig2}
\end{figure}

Changes in both time and spatial sampling values can lead to shifts in the spatial-temporal mapping relationships. As shown in Figure \ref{fig: fig2}, taking the crowd flow data from Manhattan Island, New York City, as an example, the flow data is divided into different samples for prediction $\left\{\left({Past}_{TimeID},{\ Future}_{TimeID}\right)_{ZoneID},\ TimeID=\left\{1,\ 2\right\},\ \ ZoneID=\left\{43,\ 75,\ 79\right\}\right\}$, with a time window size of 6 hours. For the same time sampling, the future trends of both samples $(Past_1, Future_1)_{43}$ and $(Past_1, Future_1)_{75}$ are similar to their historical trends, showing an overall downward trend. In contrast, the future trend of $(Past_1, Future_1)_{79}$ first follows the historical trend and then reverses. For the same region sample, the future trend of $(Past_1, Future_1)_{75}$ is opposite to its past trend, while $(Past_1, Future_1)_{75}$ shows the same future and past trends.

This difference arises from the spatial-temporal characteristics of the observation samples, i.e., the different spatial-temporal contexts, and spatial-temporal confounders can portray this information. For example, due to the different functionalities of Zone 79 compared to the other two zones, even at the same time, their ability to attract crowd flow differs, leading to different trends. While for Zone 75, the travel demand of people varies at different time windows, resulting in different trends in how the spatial region is visited across different time periods.

Therefore, we propose a Cross-Time-Attention-based Past-to-Future Mapping mechanism. To better capture the inherent spatial-temporal characteristics of STT representations, we construct Spatial-Temporal Embeddings (STE) of STTs by fusing time and spatial confounders, which preserve the global positional and structural information of the STTs in both time and space dimensions. To better capture the dynamic transformation relationship between past and future, which is a pair of cross-time entities, we introduce a Cross-Time-Attention (CTA) mechanism to enable querying between future STEs and past STEs. The queried relationships are future used to guide the transformation of past representations to future representations.

\begin{figure}[h]
	\centering
    \subfloat[]{\includegraphics[height=0.3\textwidth]{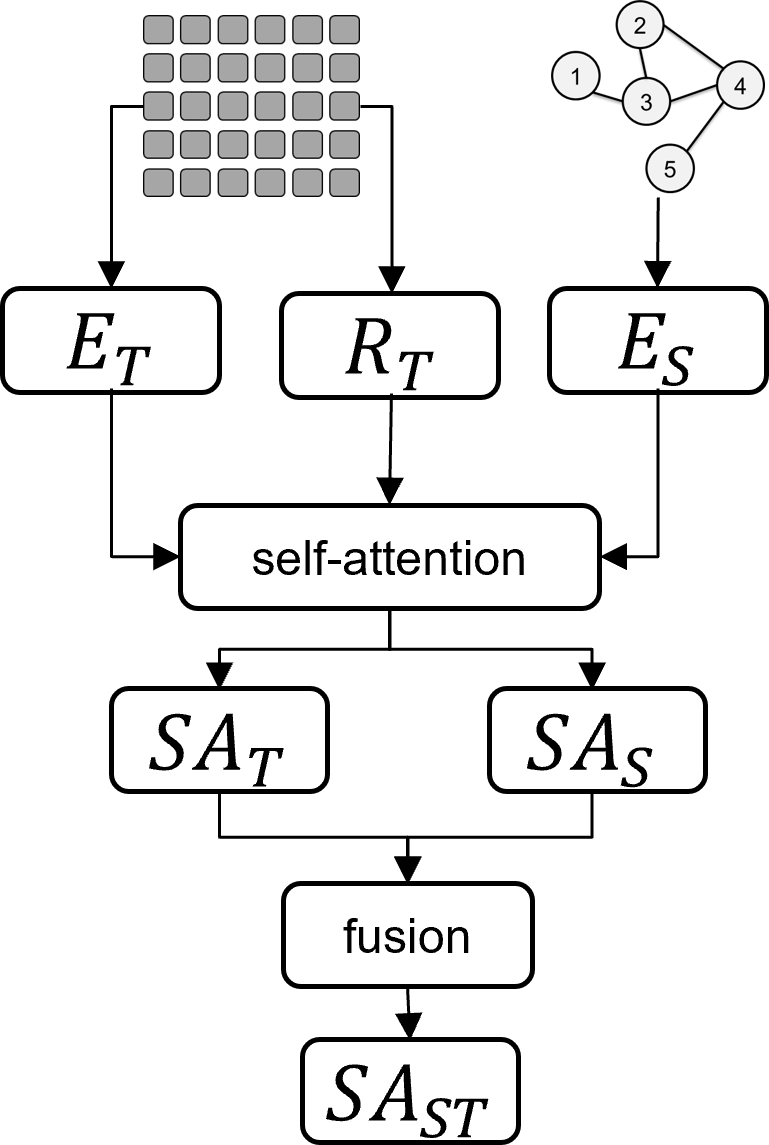}\label{fig: fig3a}}\hspace{15mm}
	\subfloat[]{\includegraphics[height=0.3\textwidth]{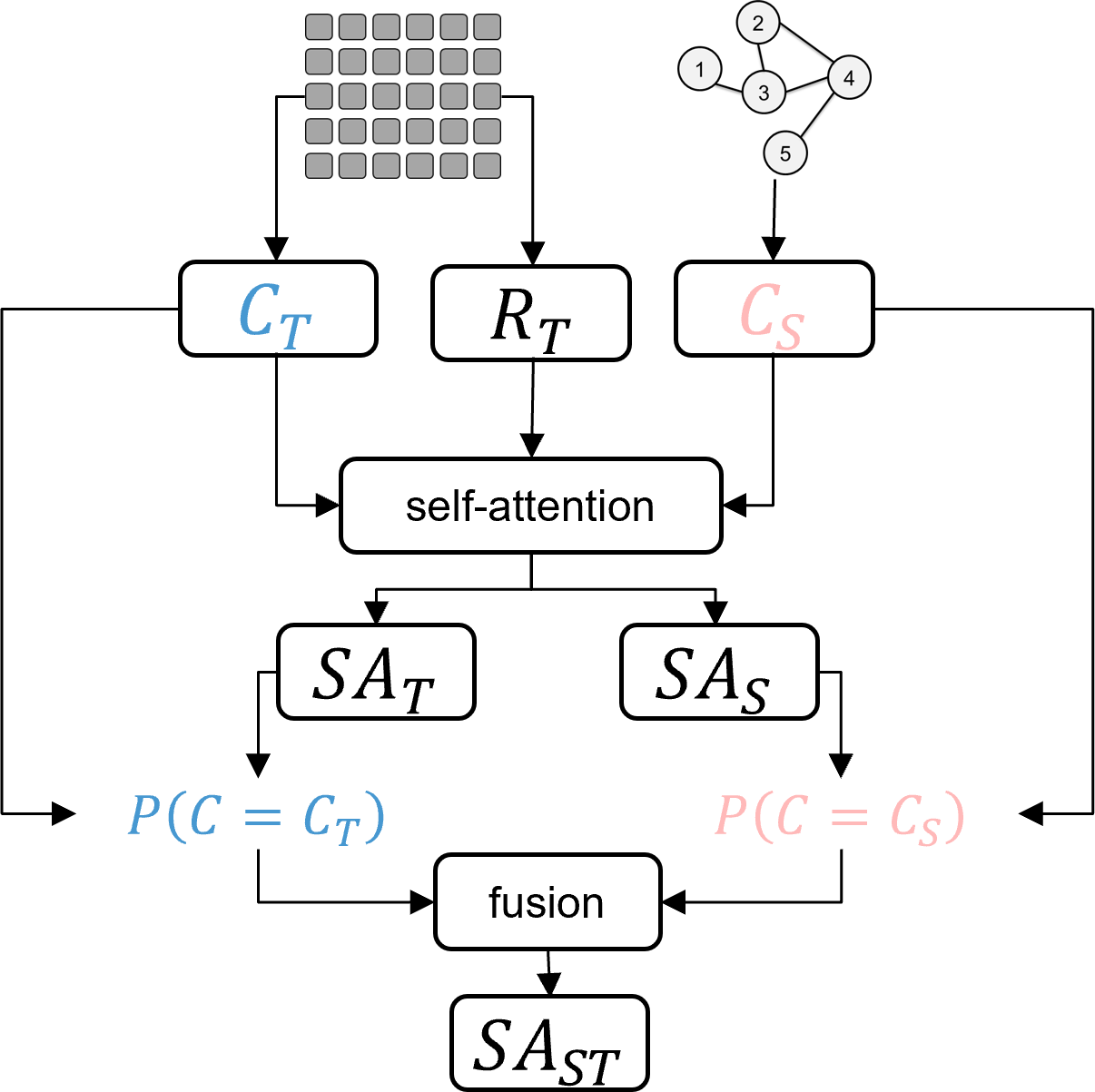}\label{fig: fig3b}}
	\caption{The fusion of STDC strategy and ST Transformer framework. (a) ST Transformer framework. (b) ST Transformer with STDC Strategy.}
\end{figure}

By combining A1 and A2, we constructed a prediction skeleton with the structure of STDC Encoder → CTA-based Past-to-Future Mapping → STDC Decoder, and instantiated this skeleton based on the Transformer architecture to obtain the STDCformer, which simultaneously achieves de-confounding and spatial-temporal prediction. Compared to existing ST Transformer architectures (e.g., Figure \ref{fig: fig3a}), STDCformer introduces confounder's information in both STE encoding and spatial-temporal representation learning (Figure \ref{fig: fig3b}), allowing the model to use confounder's information to simultaneously guide spatial-temporal mapping and spatial-temporal fusion. This approach maximizes information sharing between de-confounding representation and spatial-temporal prediction, thereby enhancing information utilization efficiency.

Our contributions are as followed:

(1) We propose a new perspective for formulating spatial-temporal prediction, and construct a novel Encoder-Past-to-Future Mapping-Decoder prediction skeleton. Based on this skeleton, a spatial-temporal prediction model, STDCformer, is developed, which achieves de-confounded crowd flow prediction.
    
(2) We introduce a spatial-temporal backdoor adjustment strategy, partitioning the confounders behind crowd flow prediction into two categories: temporal confounders and spatial confounders, and incorporating auxiliary information to accurately build confounders. The representation of confounders is used to encode spatial-temporal embeddings and guide spatial-temporal fusion, enhancing the utilization of confounders' information and simultaneously achieving de-confounding.
    
(3) We conducted prediction experiments under independent and identically distributed (IID) conditions on two real-world datasets. The experimental results show that STDCformer outperforms state-of-the-art (SOTA) models, including STGNNs and ST Transformers. Moreover, STDCformer generalizes better to out-of-distribution (OOD) datasets in a zero-shot manner. We aligned the wights of learned confounders with real-world data and discussed their physical meaning, explaining how the deconfounding strategy removes confounding bias from the data by reweighting.
    
(4) We build two crowd flow datasets for Manhattan and Brooklyn in New York City, which include auxiliary information required to build confounders for all STTs within the datasets. These datasets provide a foundation for future research on de-confounded crowd flow prediction.

\section{Related Works}

\subsection{Spatial-Temporal Graph Prediction}
Crowd Flow Prediction in Urban Areas is a typical spatial-temporal graph prediction task, and existing spatial-temporal graph prediction methods can potentially be applied to crowd flow prediction. Therefore, this paper provides a general review of spatial-temporal graph prediction methods. Spatial-temporal graph prediction aims to forecast the future states of nodes, edges, or graphs in data constructed as graphs. In such graphs, nodes typically represent spatial entities (e.g., regions), while edges represent relationships between spatial entities (e.g., adjacency, interaction). Typical spatial-temporal graph prediction tasks include traffic flow prediction, crowd flow prediction, epidemic prediction, and air quality prediction. Unlike spatial-temporal data constructed as regular grids (e.g., remote sensing images), the spatial regions in graphs are mostly irregular units, and the relationships between them are often more complex. Unlike purely time-series prediction or graph learning tasks, spatial-temporal graph prediction requires the simultaneous modeling of temporal and spatial features, and thus involves modules for capturing both temporal dependencies and spatial dependencies.

\begin{figure}[h]
\centering
\includegraphics[height=0.2\textwidth]{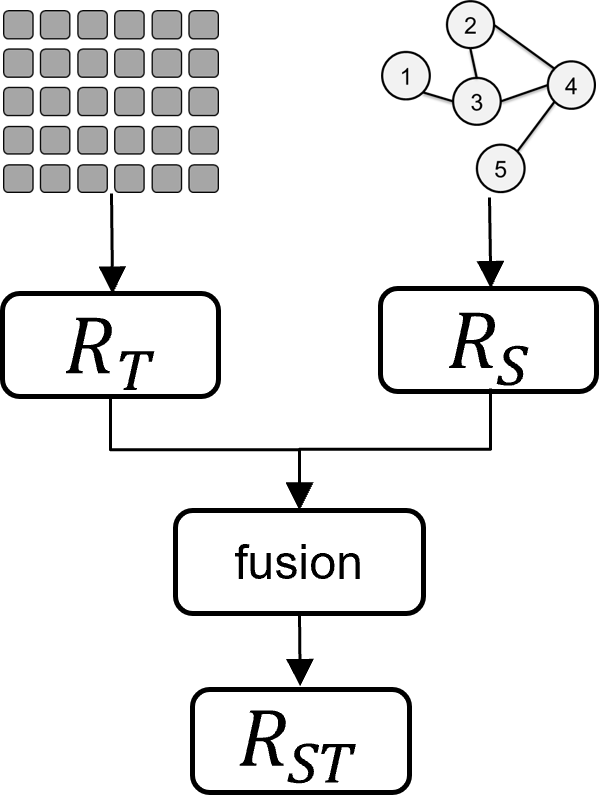}
\caption{The framework of STGNNs.}
\label{fig: fig4}
\end{figure}

Currently, mainstream spatial-temporal graph prediction frameworks can be divided into two categories: Spatial-temporal Graph Neural Networks (STGNNs) \cite{RN4,RN27,RN28} and
Spatial-temporal Transformers (ST Transformers) \cite{RN22,RN29}. As shown in Figure \ref{fig: fig4}, STGNNs utilize time-series representation models and GNNs to learn temporal representation $R_T$, and spatial representation $R_S$, then fuse these two representations to obtain the final spatial-temporal representation $R_{ST}$, for prediction. In another way, ST Transformers introduce temporal
embeddings $E_T$, and spatial embeddings $E_S$ to preserve the positions and structural information of tokens in the original spatial-temporal sequence, and use temporal self-attention and spatial self-attention modules to derive temporal representations $SA_T$ and spatial representations $SA_S$, respectively. These two representations are then fused to generate the final spatial-temporal representation for prediction (as shown in Figure \ref{fig: fig3a}).

\subsubsection{Spatial Representation}
The main goal of the spatial representation module is to model the relationships between spatial regions and encode these relationships into representations. Based on the fundamental structure of spatial representation models, existing methods can be categorized into CNN-based, GNN-based, and Transformer-based spatial-temporal prediction approaches. CNN-based models partition spatial regions into regular grids and use spatial convolutions to achieve spatial representation. For example, DeepST\cite{RN21} uses a deep neural network (DNN) to extract the spatial features of the urban grid. However, this approach is not suitable for spatial-temporal prediction in irregular regions. GNN-based spatial representation modules and Transformer-based spatial representation modules are separately the core component of existing STGNNs and ST Transformers.

\textbf{GNN-based models}. DCRNN \cite{RN6} uses a diffusion convolution to extract graph structure signals by walking randomly on the graph; STGCN\cite{RN16}, T-GCN\cite{RN10}, and KST-GCN\cite{RN7} use graph convolution networks to represent graph structures; STFGNN\cite{RN15} uses a simplified graph matrix multiplication to aggregate spatial dependencies; Graph wavenet\\cite{RN13} generalizes the form of GCN to diffusion convolution to build graph convolution layer; ASTGCRN\cite{RN5}proposes an adaptive graph convolution to learn node representations; ASTGCN\cite{RN12}uses graph attention convolution to obtain spatial representations based on spatial attention; HGCN\cite{RN14} uses spectral clustering and spatial gated graph convolution to obtain micro and macro spatial representations; MTGNN\cite{RN30} uses a mix-hop propagation layer in the graph convolution module to select and aggregate information from neighboring nodes.

\textbf{Transformer-based models}. ASTTGN\cite{RN20} leverages node2vec\cite{RN31} to retain graph structure information, and inputs it as the embedding of the node sequence into the spatial transformer model. Then, a multi-head attention mechanism is used to capture dynamic and implicit spatial dependencies. STTN\cite{RN17} inputs the representation obtained by graph convolution into the transformer structure to capture dynamic graph signals; GLSTTN \cite{RN22} constructs a spatial transformer similar to STTN to learn spatial representations; Traffic transformer\cite{RN8} constructs an encoder-decoder framework based on multi-head attention, and introduces the original graph structure as a mask in the decoder.

\subsubsection{Temporal Representation}

The primary goal of the temporal representation module is to model temporal features such as trends and periodicity in time series data. According to the fundamental structure of temporal representation models, existing methods can be categorized into RNN-based, CNN-based, and transformer-based spatial-temporal prediction models. RNN-based models primarily utilize Recurrent Neural Networks (RNNs)\cite{RN32,RN33} and their variants\cite{RN34} to extract temporal features. CNN-based models extract temporal features by applying convolutions over the time dimension. Transformer-based temporal representation models, on the other hand, use self-attention mechanisms to capture dependencies across different time steps.

\textbf{RNN-based models}. Models such as ASTGCRN\cite{RN5}, T-GCN \cite{RN10}, and KST-GCN\cite{RN7} all employ Gated Recurrent Units (GRU). DCRNN \cite{RN6} utilizes GRU to construct a Diffusion Convolutional Gated Recurrent Unit for extracting temporal features. ST-MetaNet \cite{RN11} employs a GRU as the first layer in a Seq2Seq framework. Traffic Transformer\cite{RN8}uses a simple Long Short-Term Memory (LSTM) to obtain temporal embeddings for extracting temporal features. MASTGN \cite{RN9}uses LSTM to process features after transformations based on spatial and temporal attention.

\textbf{CNN-based models}. STGCN\cite{RN16} utilizes 1-D causal convolution and Gated Linear Units (GLU) to construct temporal convolutional layers. Graph Wavenet\cite{RN13} uses Gated Temporal Causal Convolution to expand the temporal receptive field. STFGNN\cite{RN15} proposes a new gated CNN module for extracting temporal features. ASTGCN \cite{RN12} combines temporal attention with standard convolution to extract temporal features. HGCN \cite{RN14} replaces the original gated temporal convolution with dilation convolution whose dilation factor is 2 to construct temporal gated convolution.

\textbf{Transformer-based models}. ASTTGN\cite{RN20} employs an adaptive temporal transformer module to capture long-range temporal dependencies. STTN\cite{RN17} construct a temporal transformer to capture bidirectional, long-range temporal dependencies. LSTTN \cite{RN19} trains a transformer-based encoder to represent temporal features of subsequences through a self-supervised task of mask reconstruction.

\subsubsection{Spatial-Temporal Fusion}
Existing methods for spatial-temporal fusion can be divided into two categories: direct fusion and adaptive fusion. The former typically connects the temporal and spatial representation modules sequentially, interleaves them in a block or simply adding the two types of representations. In contrast, the latter involves inputting both representations into a learnable module for fusion.

\textbf{Direct Fusion}. In TGCN\cite{RN10}, the temporal representation module follows the spatial representation module, with the learned spatial representations being fed into the temporal representation module. While Graph Wavenet\cite{RN13} places the temporal module before the spatial module. G-SWaN\cite{RN35} first uses a WaveNet to obtain the temporal representation and then feeds it into the spatial graph transformer to learn the spatial representation. Similarly, Traffic Transformer\cite{RN8} uses LSTM to obtain representations, which are then input into subsequent global and local spatial graph representation modules. STGCN\cite{RN16} and HGCN \cite{RN14} build the temporal gated convolutional layer and spatial convolutional layer into a "sandwich" structure. MTGNN \cite{RN30} interleaves the temporal and spatial convolution modules for spatial-temporal fusion, while STTN \cite{RN17} interleaves the spatial transformer and temporal transformer. DCRNN\cite{RN6} replaces matrix multiplications in the GRU with spatial diffusion convolutions, using spatial representations as recurrent units to learn temporal representations. Similarly, ASTGCRN\cite{RN5} replaces the multilayer perceptron (MLP) in GRU with adaptive graph convolution. MASTGN \cite{RN9} concatenates the representations transformed by spatial attention and internal attention, using this concatenated representation as input for the followed layers.

\textbf{Adaptive Fusion}. ASTTGN\cite{RN20}, MVSTT\cite{RN36} and GMAN\cite{RN37} use a gated fusion mechanism to adaptively perform weighted fusion of temporal and spatial representations. PDFormer\cite{RN18} allocates different attention heads to the temporal and spatial representations for attention learning and then uses a multi-head attention mechanism to fuse the temporal and spatial representations.

\subsection{Spatial-Temporal De-confounding Representation Learning}
Spatial-temporal de-confounding representation learning involves modeling causal variables in spatial-temporal data, and estimating the causal effect of the treatment variable on the outcome variable by removing the influence of confounding variables (confounders). In different data and task scenarios, the confounders within the system vary, and the de-confounding methods differ accordingly. Existing research can further be categorized into methods based on structural causal model (SCM) and potential outcome framework. The underlying ideas of the two types of approaches are unified, differing only in the causal language used.

\subsubsection{Structural Causal Model based Method}
Methods based on SCM can directly implement interventions on causal graphs, transforming causal problems into statistical language that can represent the data through front-door, backdoor adjustments and other methods. Backdoor adjustment realizes de-confounding by stratifying and balancing confounding factors, being adopted by many studies. Some works categorize confounding factors into pre-defined layers. For instance, STCTN \cite{RN26} considers regional attributes in the region network as confounders, which may cause existing spatial-temporal prediction models to absorb spurious correlations between the historical and future data. To address this, spectral clustering is used to partition regional attributes, and an unbiased prediction model is constructed in an independent parameter space for each partitions to achieve de-confounded predictions. SEAD\cite{RN38} aims to eliminate the confounding effects of social environments on pedestrian trajectories. This work assumes that the confounders follow a uniform distribution, and the joint distribution of confounders and causal variables is learned through a cross-attention module. CISTGNN \cite{RN39} learns representations of confounders from external weather environments and applies backdoor criteria for de-confounding. STEVE\cite{RN25} focuses on the finiteness of the number of confounders and the completeness of their categories, dividing confounders into invariant and variant layers. CaST \cite{RN24} discretizes the confounders through a temporal environment codebook, so that the representations of confounder fall into the most appropriate layers. CTSGI\cite{RN40} leverages images at pedestrian trajectory points as the source of confounder representations, extracting environmental representations through a semantic segmentation model, and then adjusting confounding effects using backdoor criteria. Front-door adjustment, on the other hand, applies to causal graphs where no back-door paths exist, providing a de-confounding strategy different from the backdoor criterion. STNSCM \cite{RN41} uses GLUs to obtain the distribution of causal variables after intervention from historical observational data and external environment data. In this distribution, causal variables can combine with different confounders obeying an unbiased probability, allowing for de-confounded effect estimation. A similar strategy is used in vehicle trajectory prediction, where the contextual scene is treated as a confounder and eliminated through a counterfactual representation inference module based on front-door adjustment strategies \cite{RN42}. CASPER \cite{RN43} uses front-door adjustment to transform time-series completion tasks into summing over  subcategories of input data representations, and calculates de-confounded causal effects through learnable prompt vectors.

\subsubsection{Potential Outcome Framework based Method}
The main idea of methods based on the potential outcome framework is to treat different observational units as samples and simulate interventions by processing these samples comprehensively, thus obtaining de-confounded causal effects. Sample re-weighting is one of the classical methods for achieving sample balance \cite{RN44}, which is used to assign different cities and regions to different treatment groups and calculate the causal effects corresponding to each treatment. For example, it has been used in studies to assess the impact of changes in POI on regional pedestrian flow\cite{RN45}. With the development of representation learning, balancing confounders based on the obtained unified representations of the samples through representation learning can also ensure that the background attributes distribution across different experimental groups are similar. For instance, CAPE\cite{RN46} generates representations for different locations from historical event sequences and measures the differences in the representations using the Integral Probability Metric (IPM), minimizing the overall differences between samples in different experimental groups to achieve representation balance. SINet\cite{RN47} uses the Hilbert-Schmidt Independence Criterion (HSIC) as a regularizer to guide representation balance and remove the effect of hidden confounders. CIDER\cite{RN48} applies the Wasserstein distance to balance representations across different administrative regions.

In the spatial-temporal prediction module, the proposed method constructs temporal and spatial representations based on transformer models, and employs adaptive spatial-temporal fusion. In the spatial-temporal de-confounding module, the de-confounding strategy proposed in this paper is based on SCM, utilizing the backdoor criterion for de-confounding. The distinctions between this method and the aforementioned related studies are as follows: 1) The strategy employed in this paper innovatively categorizes confounders into two abstract types: temporal confounder and spatial confounder. Furthermore, the probability distributions of the confounders are learned adaptively, which avoids the pre-setting of the number of categories and the distribution of confounders; 2) The constructed spatial-temporal de-confounded representations can simultaneously participate in both spatial-temporal representation learning and spatial-temporal fusion, maximizing information sharing between the two modules.

\section{Methodology}
\subsection{Hypothesis on Spatial-Temporal Prediction}
\label{sec: hypothesis}

\begin{figure}[htbp]
	\centering
    \subfloat[]{\includegraphics[height=0.4\textwidth]{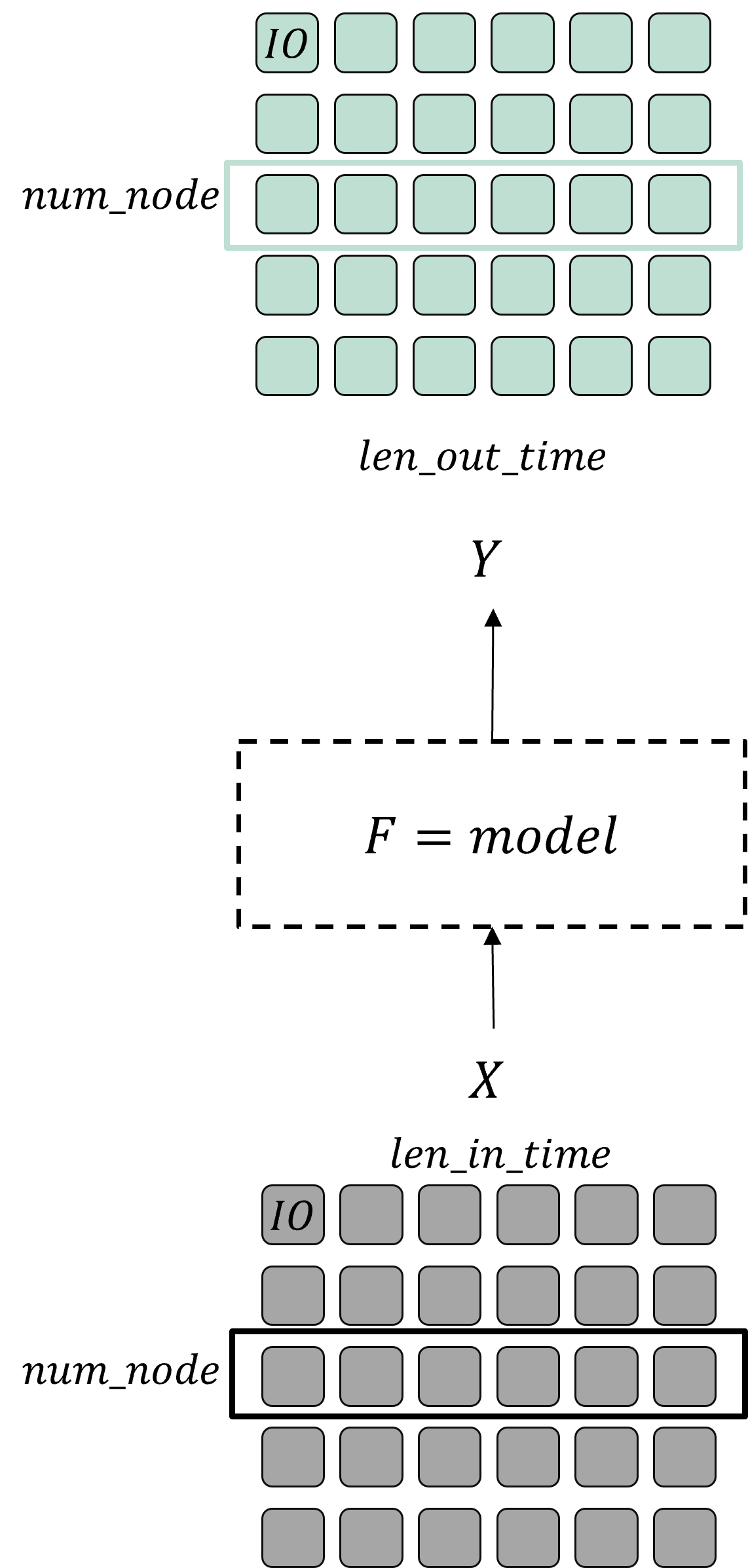}\label{fig: fig5a}}\hspace{15mm}
	\subfloat[]{\includegraphics[height=0.4\textwidth]{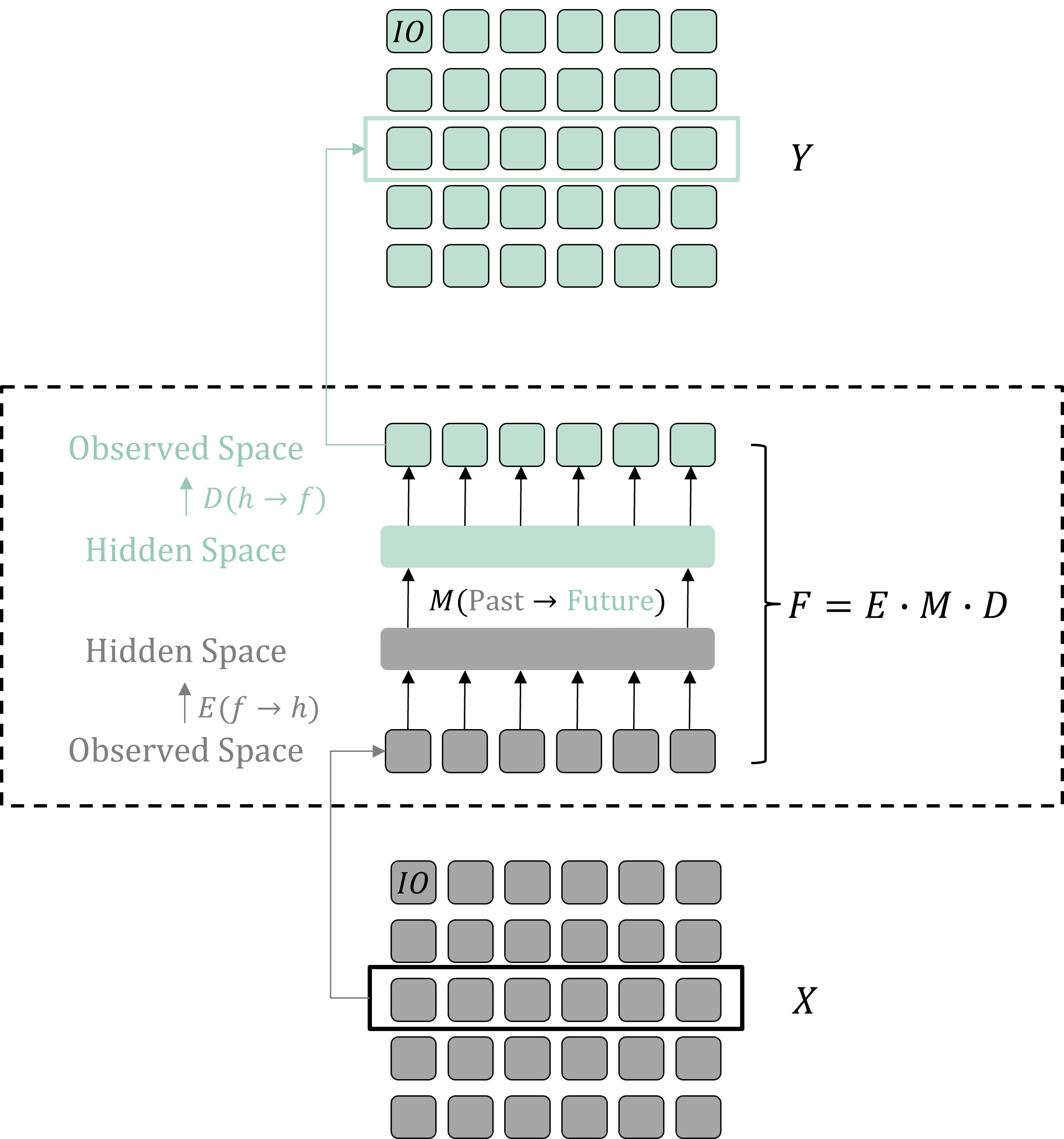}\label{fig: fig5b}}
	\caption{The formulation of crowd flow prediction. (a) Existing perspective. (b) A novel decomposing perspective proposed in this paper.}
\end{figure}

Spatial-temporal prediction aims to infer the future state for $T^f$ steps based on the known previous observations for $T^p$ steps. Taking the prediction of crowd flow as an example, the spatial-temporal prediction task is typically formalized as $X^{n \times f} \xrightarrow{F} Y^{n \times f}$, where $X^{n \times f} = [(Inflow_{t-T^p+1}, Outflow_{t-T^p+1}), ..., (Inflow_t, Outflow_t)], Y^{n \times f} = [(Inflow_{t+1}, Outflow_{t+1}), ..., (Inflow_{t+T^f}, Outflow_{t+T^f})], n$ represents the number of spatial regions to be predicted, and $f$ represents the number of features to be predicted. For the case of predicting the inflow and outflow, $f = 2$. As shown in Figure \ref{fig: fig5a}, existing methods typically use spatial-temporal prediction model to approximate $F(X \rightarrow Y)$, where the underlying assumption is that there exists some learnable pattern between past and future observations. However, in this paper, we argue that the nature of this pattern is to describe the transformation relationship between the past and future in different spaces, rather than simply viewing it as a mapping between input and output values. Therefore, as shown in Figure \ref{fig: fig5b}, we further decompose the spatial-temporal prediction task into space transformations ${E, D}$ between the observation space and the hidden representation space, and the spatial-temporal mapping $M$ from future states to past states within the representation space. The past inflow and outflow values (IO) are first mapped into a high-dimensional hidden space through a representation model $E(\cdot)$. Then, in the high-dimensional latent space, they undergo a spatial-temporal transformation $M(\cdot)$ to obtain the future IO vector, which is finally transformed back to the observable IO values through another representation model $D(\cdot)$. Therefore, the crowd flow prediction problem can be formalized as Eq. \ref{eq: 1}:

\begin{equation}
\label{eq: 1}
 IO_{t+1}, ..., IO_{t+T^f} = F(IO_{t-T^p+1}, ..., IO_t), F = E \cdot M \cdot D
\end{equation}

Based on this decomposition, the spatial-temporal prediction model must address two key questions. Q1: \textbf{What kind of representation space allows for mapping the past to the future?} Q2: \textbf{How to achieve mapping the past to the future within the representation space?} For Q1, we argue that the predictability of spatial-temporal data is ensured by the transmitting the influence of the past on the future, and this influence can be measured through causal effects. Therefore, based on causal assumptions, we propose a spatial-temporal backdoor adjustment strategy to learn a de-confounded representation space (Section \ref{sec: hypothesis1}), where past representations being encoded and future representations being decoded. For Q2, we believe that the transformation relationship between the past and the future depends on their spatial-temporal. The spatial-temporal embeddings (STE), which fuse the information of both temporal and spatial confounder can model this spatial-temporal context. Therefore, we propose a Past-to-Future Mapping Module based on a Cross-Time-Attention (CTA) mechanism, which facilitates the mapping between past and future through the query of STE from both the future and the past (Section \ref{sec: hypothesis2}).

\subsubsection{A Spatial-Temporal De-Confounded (STDC) Representation Space}
\label{sec: hypothesis1}
The foundation of the ability to infer future states from historical states for spatial-temporal prediction models is that the past has some form of "influence" on the future. This influence determines the mapping relationship between historical input data and future output data, which is the basis of the predictability of spatial-temporal data. Causal inference can measure the influence of the past on the future using causal effects, aiming to build causal relationships between variables rather than merely correlations, enabling a more accurate and stable representation of the variables\cite{RN49}. The underlying assumption is that causality is a more reliable form of association, where information flows along the causal links, transmitting the causal effect of the cause on the outcome. Based on this assumption, the spatial-temporal prediction process can be viewed as the process of estimating the causal effect of historical data representations on future data representations, inferring the outcome from the values of causes. However, the hidden confounding bias makes it impracticable to estimate the true causal effect directly from the observational data. This section takes the basic unit of spatial-temporal prediction, STT, as a starting point, to describe the causality underlying crowd flow prediction.  We also propose a spatial-temporal backdoor adjustment strategy to learn a de-confounded representation space.

\textbf{Spatial-Temporal Token (STT)}. Based on the form of input data, the basic unit in spatial-temporal data is the observation on a spatial entity $S_i$ at a specific time step $T_j$. In this paper, this smallest unit in spatial-temporal data is analogized to a token in text, referred to as the Spatial-Temporal Token (STT), represented as: $STT_{ij} = (S = S_i \in R^{1 \times s}, T = T_j \in R^{1 \times t}, V = IO_{ij} \in R^{1 \times f})$. where $V$ is the variable to be predicted, and $S$ and $T$ describe its spatial and temporal characteristics, respectively. These can be considered as background attributes of STTs, and they influence the estimation of the causal effect $V_{past} \rightarrow V_{future}$.

\begin{figure}[h]
\centering
\includegraphics[width=0.5\textwidth]{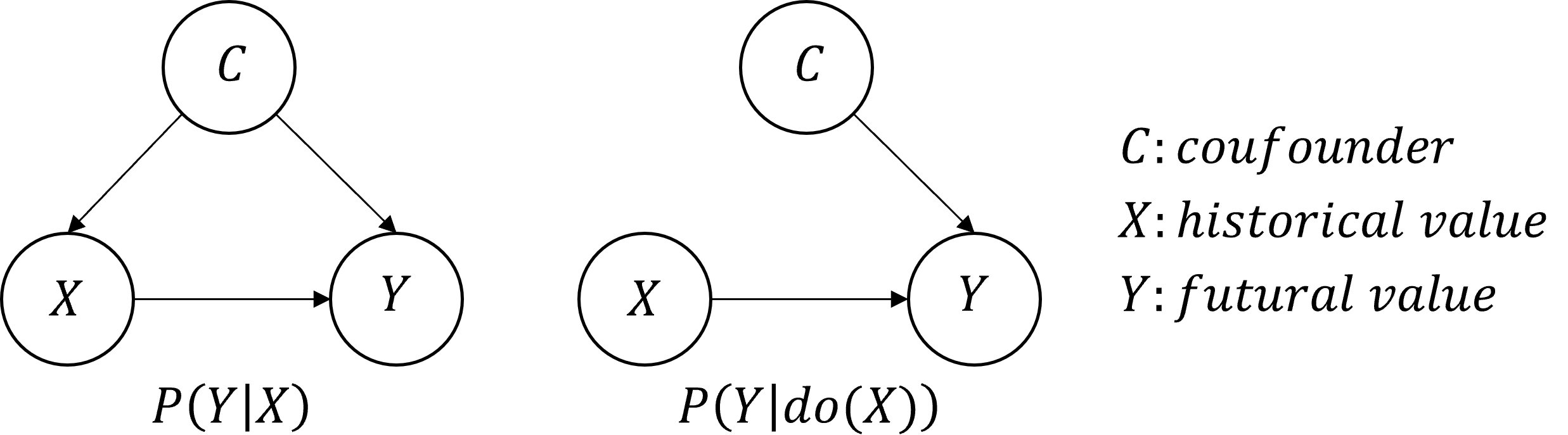}
\caption{The causal graph and underlying spatial-temporal prediction and the intervention operation, where $C = \{S, T\}$.}
\label{fig: fig6}
\end{figure}

\textbf{Causal Graph underlying Spatial-Temporal Prediction}. The prediction model infers future data representations from historical data representations, where the historical representation serves as the cause for the future representation. However, for each STT, the background attributes influence both the past and future IO representations, making $S$ and $T$ serve as confounder in the prediction process of $V$. The causal graph underlying spatial-temporal prediction can be illustrated as Fig. \ref{fig: fig6}, which suggests that the causal effect from $ V_{past} \rightarrow V_{future} $ may vary under different values of confounders. This is similar to how the effect of a drug may differ for patients of different ages in a medical trial. To estimate an accurate and fair causal effect, it is necessary to configure the trial groups in such a way that the distribution of confounders across all groups becomes uniform, i.e., to remove the confounding bias. The intervention operation on the causal graph provides an intuitive representation of de-confounding, where the arrow $C \rightarrow X$ is removed to make the distribution of $C$ independent of $X$. The causal effect after intervention is expressed as $P(Y|do(X))$.

\textbf{Spatial-Temporal Backdoor Adjustment}. Backdoor adjustment modifies the weights of observational samples with different values of confounders based on their distribution, simulating the balanced sample distribution under an intervention experiment, and is an effective way to de-confounding in observational data. It can be applied when the confounder satisfies the backdoor criterion for $X \rightarrow Y$. After backdoor adjustment, the causal effect of $X \rightarrow Y$ is expressed as: $P(Y|do(X)) = \sum_{c}^{K} P(Y|X, C = c_k) \cdot P(C = c_k)$. where the confounder is discretized into $k$ categories. However, in spatial-temporal prediction, confounders are often hidden, making it difficult to predefine the value of $k$ and the distribution of each category. To address this issue, we consider the fundamental decision factors of human movement in the real world-namely, "where" and "when"-to construct the minimal descriptive set of human movement: {"when", "where"}. We then categorize the confounders into two abstract types: temporal confounders $C_T$ and spatial confounders $C_S$. Furthermore, since STTs naturally contain the background variables $S^{1 \times s}$ and $T^{1 \times t}$, which can serve as information sources for $C_S$ and $C_T$, respectively. Based on this partition, we propose a spatial-temporal backdoor adjustment method as our de-confounding strategy, which is expressed as Eq. \ref{eq: 2}:
\begin{equation}
\label{eq: 2}
 P(Y|do(X)) = P(Y|X, C = C_S) \cdot P(C = C_S) + P(Y|X, C = C_T) \cdot P(C = C_T) 
\end{equation}

\subsubsection{A Cross-Time-Attention (CTA)-based Past-to-future Mapping}
\label{sec: hypothesis2}
The relationship between past and future states is very abstract, if using a function to approximate this kind of relationship, the parameters of this mapping function will have different value in different temporal periods. Past and future can be viewed as a pair of Cross-Time spatial-temporal entity. Capturing the transformation relationship between these entities requires addressing the following two issues:

    (1) \textbf{Representation of spatial-temporal Entities' Intrinsic Characteristics}. Based on the observations in Fig. \ref{fig: fig2}, we argue that spatial confounder and temporal confounder can jointly characterize the contextual differences between different STTs, which represent the intrinsic characteristics of tokens. Therefore, fusing the representations of temporal confounders with spatial confounder effectively captures the spatial-temporal characteristics of different spatial-temporal entities. For each STT that constitutes the Past and Future, we integrate their temporal and spatial confounder’s representations to obtain a Spatial-Temporal Embedding (STE). This embedding retains the temporal and spatial characteristics of each STT within the entire dataset, thereby representing the contextual features of different spatial-temporal entities.

    (2) \textbf{Querying Relationships Between Cross-Time Entities}. The relationships between Cross-Time entities are diverse and dynamically evolving. Therefore, attention mechanism can be effective for capturing their similarities. Since this type of querying spans across time, we refer it to Cross-Time Attention (CTA). We use the STEs of Past and Future as the Key and Query, respectively, and employ the attention mechanism to perform the query. The spatial-temporal mapping is then implemented based on the results of queried attention.

\subsection{Overview}
Based on the assumptions in Section \ref{sec: hypothesis}, we combine the physical institutions of space transformation and spatial-temporal mapping in the spatial-temporal prediction process. We implement the representation models $E(\cdot), D(\cdot)$ and the spatial-temporal mapping function $M(\cdot)$ through an Encoder-Decoder architecture and the Past-to-Future Mapping module, respectively. The prediction skeleton with architecture of \textbf{Encoder} $\rightarrow$ \textbf{Past-to-Future Mapping} $\rightarrow$ \textbf{Decoder} is built. The encoder and decoder typically share similar model structures and perform predictions in an end-to-end manner. The transformation processes involved in each part are described as follows:

\begin{enumerate}
    \item \textbf{Encoder}: The encoder projects the past data $X^{n \times f}$ into a hidden representation space, producing the intermediate representation $H_{past}^{n \times h}$. This is expressed as: $encoder: \quad H_{past}^{n \times h} = E^{f \rightarrow h}(X^{n \times f})$.
    
    \item \textbf{Past-to-Future mapping}: The spatial-temporal mapping function captures the relationship between past and future states in the hidden representation space, transforming the past representation in the hidden space into the corresponding future vector. This is expressed as: $Past-to-Future \, mapping: \quad H_{future}^{n \times h} = M^{h \rightarrow h}(H_{past}^{n \times h})$.
    
    \item \textbf{Decoder}: The decoder reconstructs the future observations $Y^{n \times f}$ from the latent representation of the future state $H_{future}^{n \times h}$, which has been transformed by the spatial-temporal mapping. This is expressed as: $decoder: \quad Y^{n \times f} = D^{h \rightarrow f}(H_{future}^{n \times h})$.
\end{enumerate}

\begin{figure}[ht]
\centering
\includegraphics[width=0.5\textwidth]{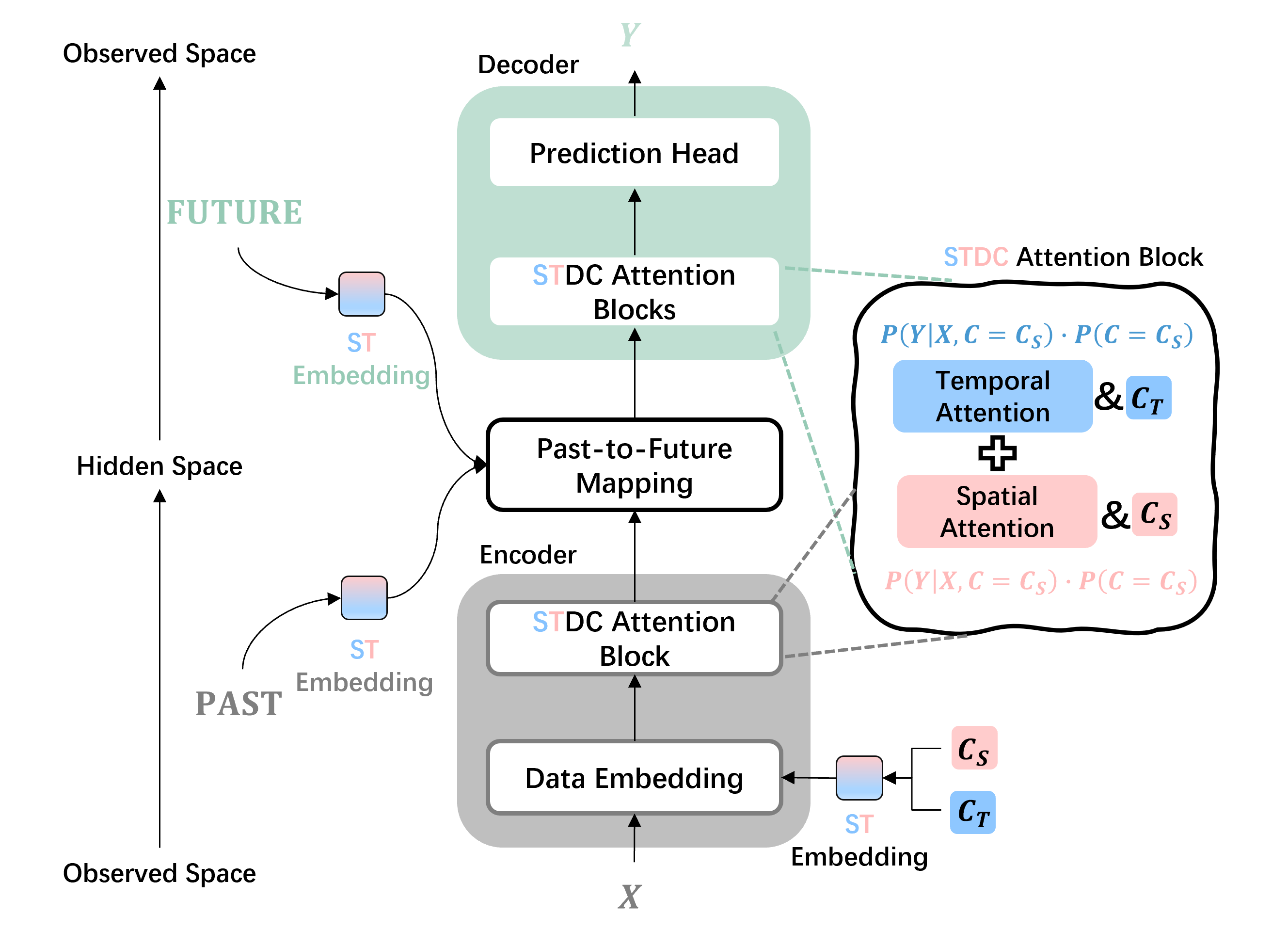}
\caption{The prediction skeleton of STDCformer which fuses the information of temporal confounders and spatial confounders.}
\label{fig: fig7}
\end{figure}

We introduce a self-attention-based transformer model to instantiate the prediction skeleton. To better achieve de-confounding in spatial-temporal prediction, we integrate the confounder information with the prediction skeleton, constructing a Spatial-Temporal De-Confounding prediction model based on Transformer (STDCformer). As shown in Figure \ref{fig: fig7}, in order to maximize the information sharing between the spatial-temporal confounders' information and the prediction skeleton, while improving the efficiency and information utilization of spatial-temporal prediction, we allow the representation of spatial-temporal confounders to participate in both space transformation and spatial-temporal mapping. This is achieved by integrating it into the prediction skeleton in the following two ways:

\begin{enumerate}
    \item \textbf{Space Transformation} $E(\cdot), D(\cdot)$: Guiding the de-confounded fusion of spatial and temporal representations. In the representation module with structures of Encoder-Decoder, the weights of temporal and spatial confounders are utilized to guide the spatial-temporal fusion process.
    \item  \textbf{Spatial-Temporal Mapping} $M(\cdot)$: Encoding Spatial-Temporal Embeddings (STE). Temporal and spatial confounders are combined to generate STEs, which are used within the Past-to-Future Mapping module to capture the spatial-temporal mapping relationships.
\end{enumerate}

\begin{figure}[h]
\centering
\includegraphics[width=0.9\textwidth]{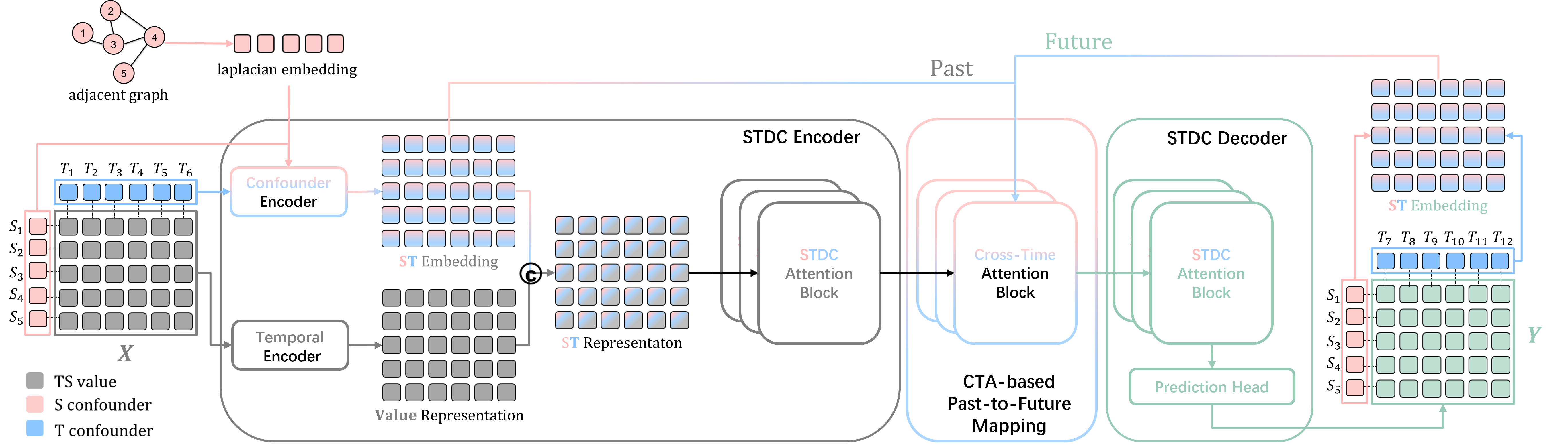}
\caption{The overall structure of STDCformer. Where the information of spatial confounders and its flow are marked as \textcolor[RGB]{255,185,185}{pink}, the information of temporal confounders and its flow are marked as \textcolor[RGB]{71,152,209}{blue}, the information of observational value and its flow are marked as \textcolor[RGB]{128,128,128}{grey}, the observational future state is marked as [RGB]{152,204,183}{green}.}
\label{fig: fig8}
\end{figure}

The STDCformer consists of the following three components, as depicted in Fig. \ref{fig: fig8}:
\begin{enumerate}
\item \textbf{STDC Encoder: Projects historical data into a Spatial-Temporal De-Confounded Representation Space}. It includes a Data Embedding Module that integrates the information of the input STTs, and a stack of Spatial-Temporal De-Confounded Attention Blocks to implement temporal representation, spatial representation, and spatial-temporal fusion.
\item \textbf{CTA-based Past-to-Future Mapping: Explicitly captures the relationship between the past and future, and performs the mapping}. It includes Cross-Time Attention Blocks to query the attention between the STEs of future and past, and further projects the past representation into the future representation.
\item \textbf{STDC Decoder: Reconstructs future observation}. It includes a stack of Spatial-Temporal De-Confounded Attention Blocks to implement temporal representation, spatial representation, and spatial-temporal fusion; and a Prediction Head that maps
\end{enumerate}

\subsection{STDC Encoder}

\subsubsection{Data Embedding}
For a $STT_{ij} = \left( S = S_i \in R^{1 \times s}, T = T_j \in R^{1 \times t}, V = IO_{ij} \in R^{1 \times f} \right)$, the data embedding layer maps it into a representation vector $\overrightarrow{STR_{ij}} \in R^{1 \times 2d}$, which is used to effectively integrate both the spatial-temporal characteristics and observational information of the STTs. The representation of the STT is constructed using the following two types of information:

\begin{enumerate}
    \item \textbf{Observational Value}. This mainly contains the observational feature of the STT. Therefore, temporal representation is used to preserve this part of the information. The flow observation value $V^{1 \times f}$ is mapped into d-dimensional space by a temporal encoder, resulting in $V' \in R^{1 \times d}$ as the value representation.
    
    \item \textbf{Spatial-Temporal Characteristics}. This mainly contains the spatial-temporal location and structural information of the STT. Since the temporal and spatial confounders representation can portray the temporal and spatial properties of the STT, a spatial-temporal embedding fusing the information of two types of confounders can preserve this part of information. A confounder encoder is used to map $T_j \in R^{1 \times t}$ and $S_i \in R^{1 \times s}$ into a d-dimensional space, yielding $C_T \in R^{1 \times d}$ and $C_S \in R^{1 \times d}$, respectively. $S_i$ is first concatenated with its representation $Laplacian_i \in R^{1 \times d_{lap}}$ in the Laplacian space of the adjacency graph, resulting in $S'_i \in R^{1 \times (s + d_{lap})}$, and then mapped. Since $C_S$ and $C_T$ capture the spatial-temporal attributes of the STT, they play a similar role to the token's embedding. Therefore, $C_S$ and $C_T$ are fused through a convolution layer to obtain the Spatial-Temporal Embedding (STE), denoted as $STE \in R^{1 \times d}$.
\end{enumerate}

Finally, the value representation $V' \in R^{1 \times d}$ and the $STE \in R^{1 \times d}$ are concatenated to form the representation $\overrightarrow{STR_{ij}} \in R^{1 \times 2d}$. The transformation within the data embedding layer is described by the following formulas in Eq. \ref{eq: 3}, where $Conv(\cdot)$ denotes the 2D convolution with the ReLU activation function.

\begin{equation}
 \label{eq: 3}
    \begin{aligned}
    V' &= Conv^{f \rightarrow d}(V^{1 \times f}) \\
    C_S &= Conv^{s + d_{lap} \rightarrow d}(S^{1 \times s} \oplus Laplacian^{1 \times d_{lap}}) \\
    C_T &= Conv^{s \rightarrow d}(T^{1 \times t}) \\
    STE &= Conv^{d \rightarrow d}(C_S^{1 \times d}) + Conv^{d \rightarrow d}(C_T^{1 \times d}) \\
    \overrightarrow{STR} &= V'^d + STE^d
    \end{aligned}
\end{equation}

\subsubsection{Spatial-Temporal De-Confounded Attention Block}
For the spatial-temporal representation within the historical time window, the observation time steps are defined as $T = (t - T^p + 1, \ldots, t)$, and the spatial-temporal representation of the region $S_i$ is given by $STR_{S_j} = \left( \overrightarrow{STT_{1T_j}}, \ldots, \overrightarrow{STT_{nT_j}} \right) \in R^{n \times 2d}$. For the observed spatial region $S = (1, \ldots, n)$, the spatial-temporal representation at time step $T_j$ is given by $STR_{S_j} = \left( \overrightarrow{STT_{1T_j}}, \ldots, \overrightarrow{STT_{nT_j}} \right) \in R^{n \times 2d}$. Therefore, the spatial-temporal representation of the data within the prediction window is for all regions on all steps and is given by $STR_{ST} = \left( \overrightarrow{STT_{1T}}, \ldots, \overrightarrow{STT_{nT}} \right) \in R^{T^p \times n \times 2d}$. It is then fed into the first layer of the Spatial-Temporal De-Confounded Attention Block in the STDC Encoder.

\begin{figure}[h]
\centering
\includegraphics[width=0.4\textwidth]{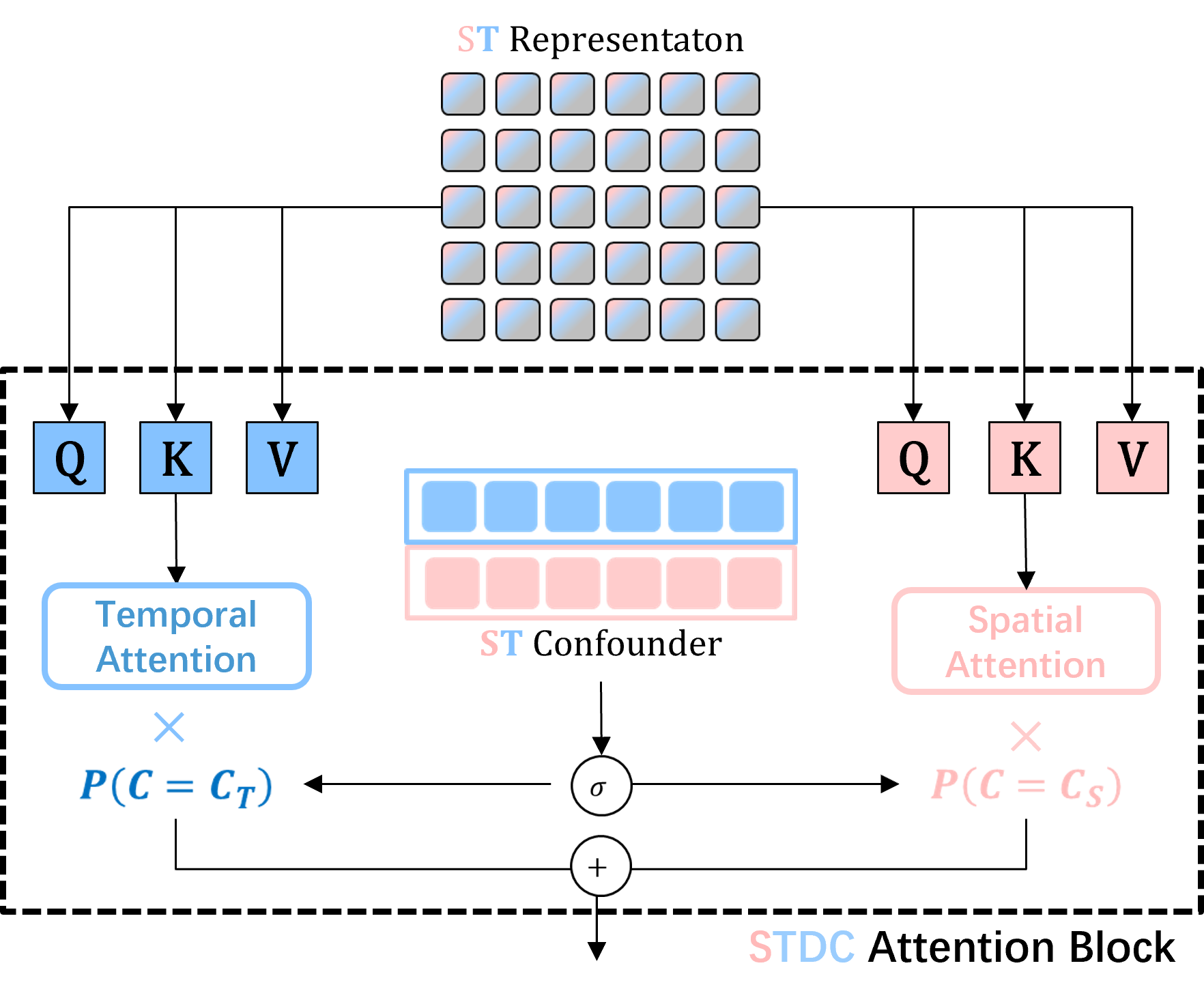}
\caption{ The architecture of Spatial-temporal De-Confounded attention block.}
\label{fig: fig9}
\end{figure}

As illustrated in Fig. \ref{fig: fig9}, the Spatial-Temporal De-Confounded Attention Block is composed of three modules: Spatial Attention, Temporal Attention, and Spatial-Temporal De-Confounding Fusion, which are designed to capture spatial representations, temporal representations, and spatial-temporal fusion, respectively. Spatial Attention and Temporal Attention use the self-attention mechanism to capture spatial and temporal dependencies between different STTs. Spatial-Temporal De-Confounding Fusion applies a spatial-temporal backdoor adjustment strategy to control spatial-temporal confounders, ensuring that the predictions are not influenced by confounding biases.

\textbf{Spatial attention.} The spatial attention module is designed to model the spatial dependencies between different regions within a given time window. In this module, the self-attention mechanism is employed to capture the similarities between STTs distributed across different spatial regions, facilitating the fusion of information across space. For each batch of data, $STR_{ST_j} = STR_{ST}[T_j, :, :] \in R^{n \times 2d}$ is used to learn the spatial attention distribution for time step $T_j$. $Q_S^{T_j}, K_S^{T_j}, V_S^{T_j} \in R^{T_P \times d}$ are firstly learned for current input representation to learn the representations of Query, Key and Value. The spatial attention coefficient matrix $A_S^{T_j} \in R^{n \times n}$ is then computed. Formulas for this process are in Eq. \ref{eq: 4}:

\begin{equation}
 \label{eq: 4}
    \begin{aligned}
    Q_S^{T_j} &= STR_{ST_j} W_{ST_j}^Q \\
    K_S^{T_j} &= STR_{ST_j} W_{ST_j}^K \\
    V_S^{T_j} &= STR_{ST_j} W_{ST_j}^V \\
    A_S^{T_j} &= \frac{Q_S^{T_j} \cdot \left(K_S^{T_j}\right)^T}{\sqrt{d}}
    \end{aligned}
\end{equation}

Where $W_{ST_j}^Q, W_{ST_j}^K, W_{ST_j}^V \in R^{2d \times d}$. Finally, for time step $T_j$, the data representation after the spatial attention transformation is given by $STR_{ST_j}' = \text{softmax}\left(A_S^{T_j}\right) \cdot V_S^{T_j} \in R^{n \times d}$. For the entire input time window, the representation is $STR_S' = \text{softmax}\left(A_S^{T_j}\right) \cdot V_S^{T_j} \in R^{T_P \times n \times d}$.

\textbf{Temporal attention.} Temporal attention module is designed to model the temporal dependencies of a specific region across different time steps. In this module, the self-attention mechanism is employed to capture the similarities between STTs distributed across different time steps, facilitating the fusion of information across time. For each batch of data, $STR_{S_iT} = STR_{ST}[:, S_i, :] \in R^{T_P \times 2d}$ is used to learn the temporal attention distribution for region $S_i$. $Q_{S_i}^T, K_{S_i}^T, V_{S_i}^T \in R^{T_P \times d}$ are firstly learned for current input representation to learn the representations of Query, Key and Value. The temporal attention coefficient matrix $A_S^{T_j} \in R^{T_P \times T_P}$ is then computed. Formulas for this process are in Eq. \ref{eq: 5}:

\begin{equation}
 \label{eq: 5}
\begin{aligned}
Q_{S_i}^T &= STR_{S_iT} W_{S_iT}^Q \\
K_{S_i}^T &= STR_{S_iT} W_{S_iT}^K \\
V_{S_i}^T &= STR_{S_iT} W_{S_iT}^V \\
A_{S_i}^T &= \frac{Q_{S_i}^T \cdot \left(K_{S_i}^T\right)^T}{\sqrt{d}}
\end{aligned}
\end{equation}

Where $W_{S_iT}^Q, W_{S_iT}^K, W_{S_iT}^V \in R^{2d \times d}$. Finally, for time step $T_j$, the data representation after the temporal attention transformation is $STR_{S_iT}' = \text{softmax}\left(A_{S_i}^T\right) \cdot V_{S_i}^T \in R^{T_P \times d}$. The representation for the entire input time window is $STR_T' = \text{softmax}\left(A_{S_i}^T\right) \cdot V_{S_i}^T \in R^{T_P \times n \times d}$.

\textbf{Spatial-Temporal De-Confounded Fusion.} The Spatial-Temporal De-Confounded Fusion module fuses the spatial and temporal representations obtained from the previous two modules with simultaneously controlling the confounders. According to the spatial-temporal backdoor adjustment strategy $P(Y|do(X)) = P(Y|X, C = C_S) \cdot P(C = C_S) + P(Y|X, C = C_T) \cdot P(C = C_T)$, $P(X, C = C_S)$ here, $P(X, C = C_S)$ represents the joint distribution of spatial confounders and spatial representations, while $P(X, C = C_T)$ represents the joint distribution of temporal confounders and temporal representations. Since $\overrightarrow{STT_{ij}}$ contains both $C_{S_i}$ and $C_{T_j}$, which are involved in the computation of spatial attention and temporal attention, respectively, $STR_S'$ and $STR_T'$ can be considered as the representations of $P(X, C = C_S)$ and $P(X, C = C_T)$. For each $STT_{ij}$, to obtain $P(C = C_S)$ and $P(C = C_T)$, $C_{S_i}$ and $C_{T_j}$ are summed and passed through a Sigmoid function, mapping the weight of the two confounding factors to the range from 0 to 1. Finally, after fusion, the final spatial-temporal representation $H \in R^{T_P \times n \times d}$ is obtained for all STTs in the input data, as shown by Eq. \ref{eq: 6}:

\begin{equation}
\label{eq: 6}   
    \begin{aligned}
    & P(C_S) = \text{Sigmoid}(C_S + C_T) \\
    & P(C_T) = 1 - P(C = C_S) \\
    & H = P(C_S) \cdot STR_S' + P(C_T) \cdot STR_T'
    \end{aligned}
\end{equation}

\subsection{CTA-based Past-to-Future Mapping}
\begin{figure}[h]
\centering
\includegraphics[width=0.4\textwidth]{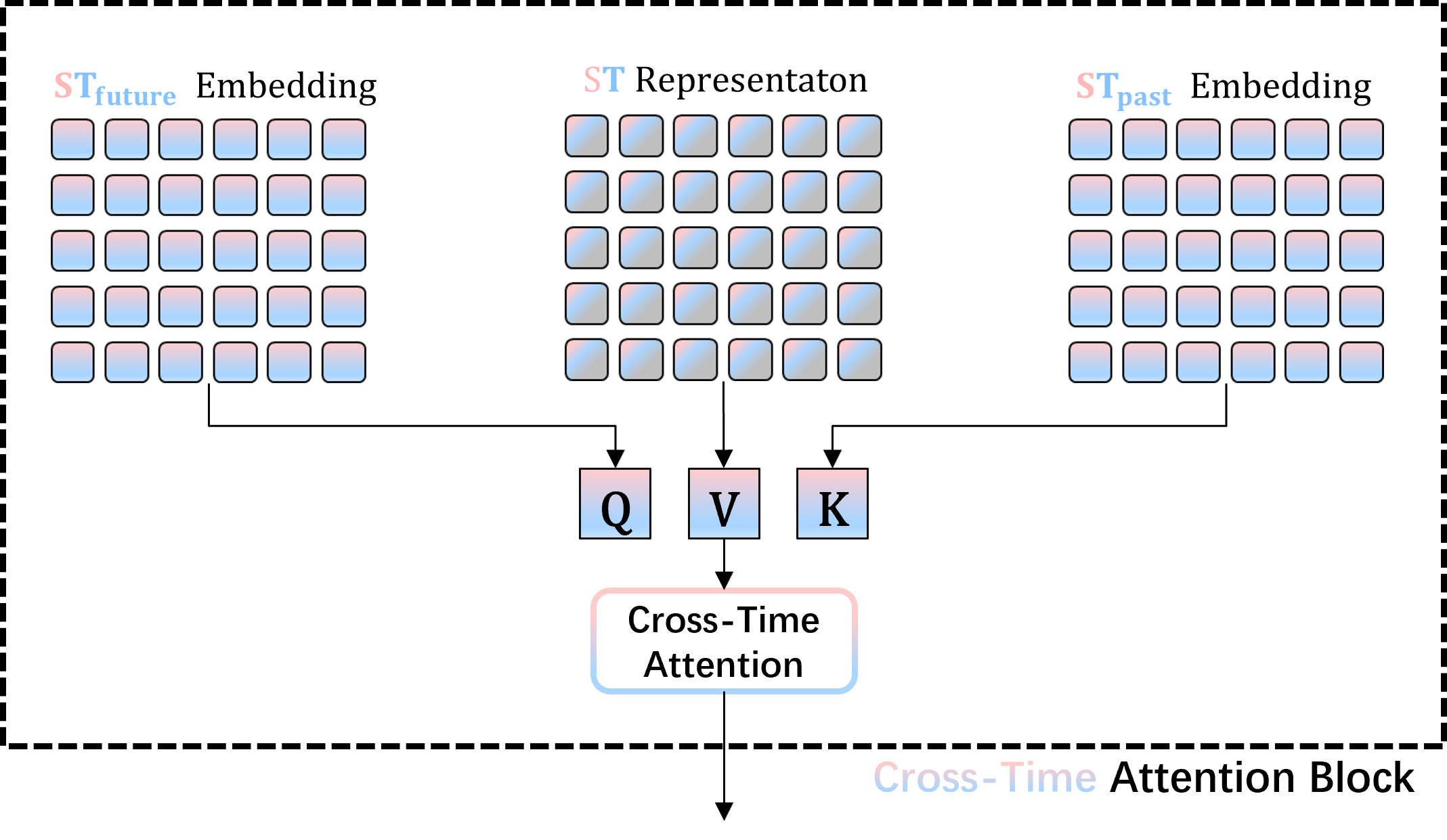}
\caption{The architecture of Cross-Time attention block.}
\label{fig: fig10}
\end{figure}

This module introduces a Cross-Time-Attention-based Past-to-Future Mapping mechanism, implemented by the Cross-Time Attention block (as shown in Fig. \ref{fig: fig10}), aimed at capturing the mapping relationship between the historical and future states in the representation space. This relationship is not dependent on the specific observational values, but rather determined by the spatial-temporal characteristics of the STTs before and after the mapping. Therefore, the Cross-Time Attention block captures the relationship between the STE of the historical window $STE_{past}$ and the future window $STE_{future}$, and based on this relationship, performs the transformation from the historical vector $H_{past}$ to the future vector $H_{future}$. To capture this complex and dynamic relationship, we utilize the attention mechanism to compute the similarity between the spatial-temporal characteristics of $STE_{past}$ and $STE_{past}$. Since attention is used to query the cross-time relationship between the STEs of the past and future representations, this process is referred to as Cross-Time Attention. We use the STE of the STTs to obtain the Query and Key, and use $H_{past}$ to obtain the Value, resulting in $H_{future}$. The formula for this process is in Eq. \ref{eq: 7}:

\begin{equation}
\label{eq: 7}   
\begin{aligned}
Q^{\text{MAP}} &= STE_{future} W_{\text{MAP}}^Q \\
K^{\text{MAP}} &= STE_{past} W_{\text{MAP}}^K \\
V^{\text{MAP}} &= H_{past} W_{\text{MAP}}^V \\
A^{\text{MAP}} &= \frac{Q^{\text{MAP}} \cdot (K^{\text{MAP}})^T}{\sqrt{d}}
\end{aligned}
\end{equation}

Where $W_{S_iT}^Q, W_{S_iT}^K, W_{S_iT}^V \in R^{d \times d}, A^{\text{MAP}} \in R^{n \times T_P \times T_f}$. Finally, the data representation after the mapping attention transformation is obtained as $H_{future} = \text{softmax}(A^{\text{MAP}}) \cdot V^{\text{MAP}} \in R^{T_f \times n \times d}$.

\subsection{STDC Decoder}
\subsubsection{Spatial-Temporal De-Confounded Attention Block}
The STDC Decoder and STDC Encoder share the same structure of the Spatial-temporal De-Confounded attention block. The key difference lies in the input: the STDC Encoder takes the historical data's ST representation as input, while the Decoder takes the future state ST representation after Past-to-Future mapping. Therefore, $STR_{ST'} = (\overrightarrow{STT_{1T'}}, ..., \overrightarrow{STT_{nT'}}) \in R^{T_f \times n \times 2d}$ is fed into the first layer of the Spatial-temporal De-Confounded attention block in the Decoder, where $T' = (t + 1, ..., t + T_f)$.

\subsubsection{Prediction Head}
The Prediction Head maps the de-confounded representations of the future state from the Spatial-temporal De-Confounded attention block to the observational feature space. In this paper, a simple convolutional layer is used to map the vector $H_{future} \in R^{T_f \times n \times d}$ in the high-dimensional representation space to the observation space. It is formulated as $\hat{Y} = Conv2d(H_{future}) \in R^{T_f \times n \times f}$. This directly implements multi-step forecasting for $T_f$ steps. The model is trained end-to-end with the Mean Absolute Error (MAE) as the loss function. The loss function is in Eq. \ref{eq: 8}:

\begin{equation}
\label{eq: 8}
l = \frac{1}{T_f} \sum_{1}^{T_f} |Y_t - \hat{Y}_t|
\end{equation}
Where $Y_t$ represents the ground truth.

\section{Experiments Results and Analysis}

\subsection{Databases}
\begin{figure}[h]
\centering
\includegraphics[width=0.3\textwidth]{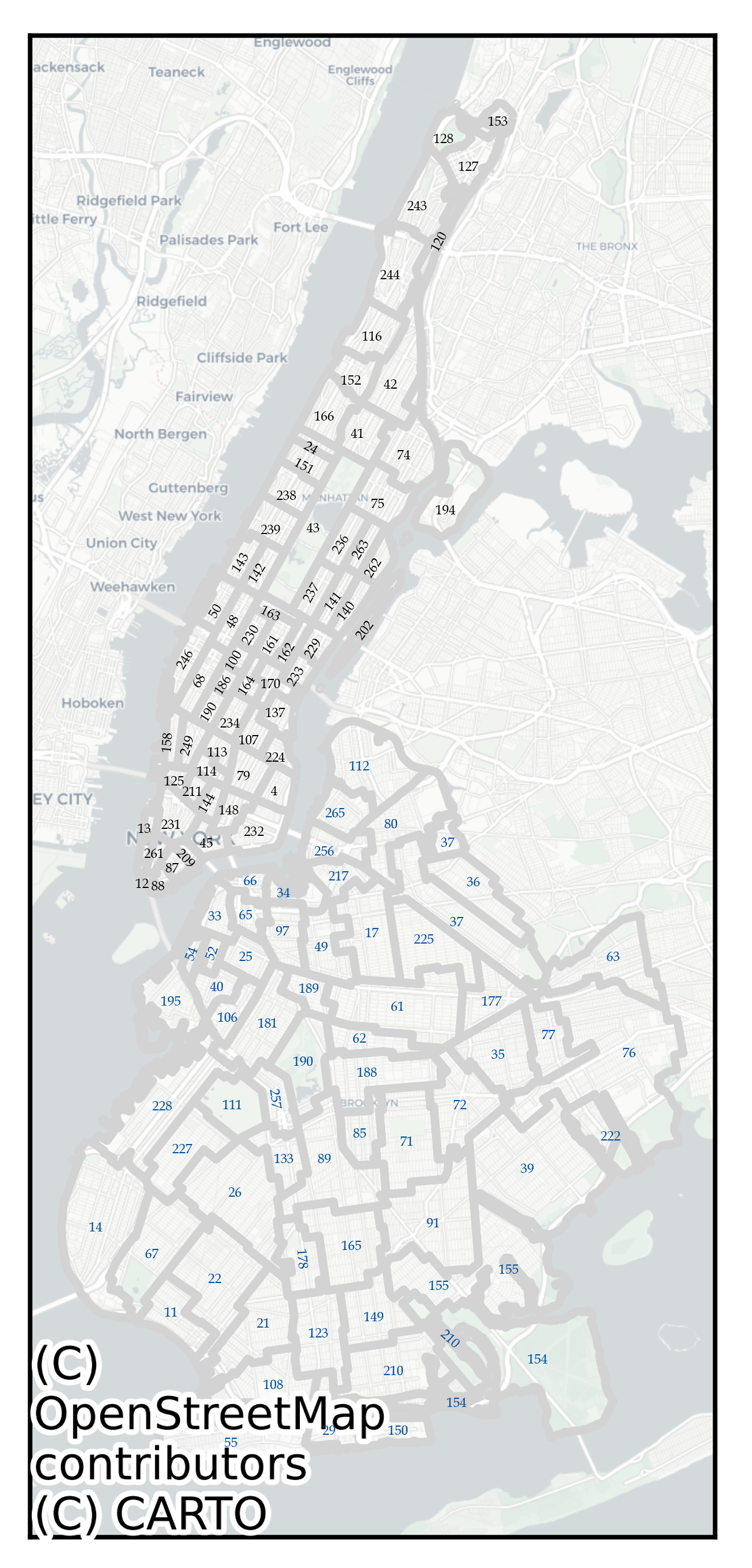}
\caption{Study area.}
\label{fig: fig11}
\end{figure}

\textbf{Crowd flow dataset}. This study utilizes two real-world crowd flow datasets, respectively for the Manhattan (MHT) and Brooklyn (BKL) areas in New York City. These datasets are derived from the New York City Taxi and Limousine Commission (TLC), which provides data on taxi passengers’ drop-in and drop-off activities. Each dataset is partitioned based on the irregular regions defined by TLC’s taxi zone. For consistency, the ID of regions used in this paper follow TLC's official zoning system, with taxi zones that are not accessible to taxis being removed. The study areas are shown in Figure \ref{fig: fig11}, with taxi zones in the MHT dataset and BKL dataset marked in black and blue, respectively. In the dataset, for each region at a given time step, inflow data represents the number of passengers dropping off in that region during the specified time window while outflow data represents the number of passengers dropping in in that region during the same time window. These inflow and outflow data are aggregated from three main taxi companies operating in the region: Yellow, Green, and For-Hire Vehicles. The statistical information for these datasets is summarized in Table \ref{tab: tab1}:

\begin{table}[]
    \centering
    \begin{tabular}{ccccc}
        \toprule
        Dataset & Number of Regions& Duration & Interval & Timesteps \\
        \midrule
        MHT & 66 & 2023/11/1-2024/6/29 & 1H & 5808 \\
        BKL & 61 & 2023/11/1-2024/6/29 & 1H & 5808 \\
        \bottomrule
    \end{tabular}
    \caption{ The basic information of datasets.}
    \label{tab: tab1}
\end{table}

\textbf{Auxiliary data}. To learn the confounders’ representations of STTs, this study collected auxiliary data to characterize the background characteristics of the STTs. To comprehensively and accurately introduce auxiliary information that captures the key semantic concepts influencing human movement, the information that depicts temporal and spatial confounder was collected for each STT. The auxiliary information introduced and their corresponding semantic concepts are listed in Table \ref{tab: tab2}.
\begin{enumerate}
    \item \textbf{Temporal confounders}. For each time step, we constructed a temporal confounder vector to characterize attributes such as travel necessity (e.g., whether in the human preferred movement period) and travel suitability at that specific time. The temporal confounder's information includes: time\_in\_day, day\_in\_week, whether it is a holiday, temperature and weather conditions. Weather data were sourced from wunderground.com.
    \item \textbf{Spatial confounders}. For each region, we constructed a spatial confounding vector to capture attributes such as geographic location, functionality, travel cost, and travel safety. The spatial confounder’s information includes: taxi zone ID, the number and categories of POI, the number of houses for sale and their average price, the number of reported crime cases. The taxi zone IDs are sourced from the official TLC numbering system. Housing data is obtained from realtor.com. POI data is collected from Google Maps. Crime statistics are sourced from the New York City Police Department (NYPD) official website.
\end{enumerate}

\begin{table}[]
    \centering
    \begin{tabular}{lll}
    \toprule
    Category of confounder 
    & Semantic concepts
    & Information  \\
    \midrule
    \multirow{2}{*}{Temporal confounder} 
    & travel necessity
    & time\_in\_day,   day\_in\_week, is holiday or not  \\
    & travel suitability  & weather condition, temperature   \\
    \midrule
    \multirow{4}{*}{Spatial confounder}  
    & geographic location                             
    & taxi zone ID, Laplacian eigenvalues of adjacency graph \\
    & functionality                                                                       & category of POI, number of each POI category  \\
    & habitability                                                                        & number of house for sale, average house price \\
    & movement safety                                                                   
    & shooting cases, complaint cases of citizen during the last year.\\
    \bottomrule
    \end{tabular}
    \caption{Auxiliary information used to build confounders and the taxonomy.}
    \label{tab: tab2}
\end{table}

\subsection{Baselines and Metrics}
\textbf{Baselines}. This paper selects three types of spatial-temporal prediction models as baselines: traditional machine learning models, STGNNs, and ST Transformers. Specifically, the traditional machine learning models include GRU, LSTM, and RNN, which represent fully-connected GRU, LSTM, and RNN, respectively. The STGNNs consist of T-GCN \cite{RN10}, STGCN \cite{RN16}, HGCN\cite{RN14}, Graph WaveNet (GWNET) \cite{RN13}, DCRNN\cite{RN6}, and MTGNN\cite{RN30}. The ST Transformers include GMAN\cite{RN37}, STTN\cite{RN17}, and PDFormer \cite{RN18}.

\textbf{Metric}. Mean Absolute Error (MAE) and Root Mean Squared Error (RMSE) are used as evaluation metrics to measure the difference between the actual values and the predicted values. These two metrics are commonly used for regression tasks, with MAE providing a straightforward measure of prediction accuracy, and RMSE giving more weight to larger errors, which is helpful for evaluating models where large errors are particularly problematic. All models are trained and inferred on the same device, using the same data partitioning and preprocessing steps to ensure fairness in evaluation.

\subsection{Experiments Setup}
To evaluate the effectiveness of the proposed model, we assess its IID prediction capability and OOD generalization ability after integrating the de-confounding strategy. The performance experiments are conducted in two aspects: prediction experiments under IID conditions and zero-shot transfer experiments under OOD conditions. The experimental settings are detailed below.

\textbf{General settings}. For task setup, the crowd flow prediction task is set to predict the next 6 time steps (6 hours) based on the previous 6 time steps. The final results are presented as multi-step forecasting. For model optimization, STDCformer is optimized using the Adam optimizer with an initial learning rate of 0.001. The learning rate is automatically adjusted based on the model's performance on the validation set. The training process is set to 120 epochs, with an early stopping strategy applied after 50 steps. All baseline models follow their original training parameter settings. For hyperparameter settings, for the STDCformer model, the feature dimension of the hidden representation space is set to 64, and the dimension of the Laplacian eigenvalues is set to 8. The number of ST Attention blocks is set to 5 for the MHT dataset and 6 for the BKL dataset. The attention mechanism uses 8 attention heads stacked. For model implementation, all models are trained on an NVIDIA A100 with 1 Core GPU and 64GB of memory.

\textbf{Prediction experiments under IID conditions}. Traditional crowd flow prediction tasks are conducted on the same dataset across training, validation, and testing phases, under the assumption that all samples in the dataset follow an IID distribution. In the IID prediction experiments, the training, validation, and testing data are from the same dataset, with a split ratio of 7:1:2. The training and validation sets are randomly shuffled and standardized using a standard scaler before being fed into the model. The batch size is set to 64. The model is trained on the training set, tuned on the validation set, and the best-performing model is selected to be tested on the test set. Considering that inflow and outflow represent two distinct features, capturing the attraction and repulsion of a region to human crowds during a given time period, they reflect different semantic meanings. Therefore, the model's predictions are evaluated separately for Inflow, Outflow, and their combined performance (IO).

\textbf{Zero-shot transfer experiments under OOD conditions}. The crowd flow varies significantly across different regions, and the distribution of observations differs both temporally and spatially. Therefore, the transfer of models across datasets from different regions can be viewed as a Spatial OOD problem. Spatial OOD generalization is particularly challenging for spatial-temporal graph predictions because the spatial graphs corresponding to different regions have different sizes. Most spatial dependency modeling modules in existing models cannot directly scale to another graph in a zero-shot manner. In contrast to graph convolution, self-attention in the spatial axes is not constrained by the original graph size and is well-suited to solve this problem. Therefore, we choose to compare the zero-shot spatial OOD transfer capability of two SOTA spatial-temporal transformers that use self-attention for spatial aggregation as baselines. Given that the prediction difficulty of the BKL dataset is lower and more appropriate for the zero-shot test scenario, we directly use the best model trained on the MHT dataset to test on the BKL dataset.

\subsection{Comparison experiments}

\subsubsection{Prediction Experiments Under IID Condition}
\label{sec: iid}
In order to evaluate the model's predictive ability for different features and the overall flow, we test the model's predictions for Inflow, Outflow, and the combined Inflow-Outflow (IO). The results are shown in Table \ref{tab: tab3}.

For different datasets, the model's performance on the BKL dataset is superior to that on the MHT dataset, with smaller differences in prediction metrics. This is due to the fact that the MHT dataset exhibits an overall larger range of numerical variations. As the most famous and active central district of New York City, the complexity of crowd movement in MHT is higher, making prediction more challenging. For different model architectures, ST Transformers outperform most of the STGNNs and traditional models, indicating that the introduction of self-attention mechanisms is effective in capturing complex spatial-temporal dependencies. However, powerful STGNN models such as DCRNN and MTGNN demonstrate comparable or even stronger predictive performance than some ST Transformers. This suggests that STGNNs still have significant advantages in modeling temporal and spatial dependencies and, in certain cases, can achieve accurate predictions with lower cost.

Since the model is trained based on the overall crowd flow characteristics (IO), IO can be considered the primary evaluation metric. In terms of overall feature prediction, the proposed STDCformer achieves the highest IO prediction accuracy across both datasets, while PDFormer demonstrates the second-best performance. MTGNN, GMAN, and DCRNN show comparable performance to PDFormer. Regarding the prediction accuracy of individual features, STDCformer and PDFormer remain highly competitive. PDFormer outperforms our model in inflow prediction on the MHT dataset, but our model performs better in outflow prediction. This indicates that different models may focus on certain data features during training, influenced by the characteristics of the data. On one hand, it suggests that a key direction for future research is understanding how models balance their learning capacity for different features. On the other hand, it also implies that relevant application departments can select models based on specific strengths, depending on their practical needs.

\begin{table}[h]
\centering
\resizebox{\linewidth}{!}
{\begin{tabular}{cccccccccccccc}
    \toprule
    \multicolumn{2}{c}{\multirow{3}{*}{\textbf{Model}}} & \multicolumn{6}{c}{\textbf{MHT}} & \multicolumn{6}{c}{\textbf{BKL}} \\ 
    \cline{3-14} 
    \multicolumn{2}{c}{} & \multicolumn{3}{c}{\textbf{MAE}} & \multicolumn{3}{c}{\textbf{RMSE}} & \multicolumn{3}{c}{\textbf{MAE}} & \multicolumn{3}{c}{\textbf{RMSE}} \\ 
    \cline{3-14} 
    \multicolumn{2}{c}{} & \textbf{In} & \textbf{Out} & \textbf{IO} & \textbf{In} & \textbf{Out} & \textbf{IO} & \textbf{In} & \textbf{Out} & \textbf{IO} & \textbf{In} & \textbf{Out} & \textbf{IO} \\ 
    \midrule
    \multirow{3}{*}{Traditional} & GRU & 15.21 & 17.07 & 16.12 & 25.02 & 28.52 & 26.83 & 3.64 & 3.11 & 3.38 & 5.20 & 4.63 & 4.92 \\ 
     & LSTM & 15.59 & 17.63 & 16.58 & 25.69 & 29.59 & 27.71 & 3.6 & 3.09 & 3.35 & 5.20 & 4.64 & 4.93 \\ 
     & RNN & 15.65 & 17.87 & 16.73 & 25.96 & 29.72 & 27.9 & 3.66 & 3.19 & 3.43 & 5.32 & 4.81 & 5.07 \\ 
    \midrule
    \multirow{6}{*}{STGNN} & T-GCN & 21.04 & 22.99 & 21.99 & 33.34 & 37.02 & 35.23 & 4.28 & 3.58 & 3.93 & 6.13 & 5.23 & 5.70 \\ 
     & STGCN & 16.92 & 19.10 & 17.98 & 27.44 & 31.78 & 29.69 & 3.8 & 3.25 & 3.53 & 5.44 & 4.77 & 5.12 \\ 
     & HGCN & 23.06 & 25.34 & 24.17 & 36.26 & 40.95 & 38.68 & 4.51 & 3.65 & 4.09 & 6.63 & 5.16 & 5.94 \\ 
     & GWNET & 16.38 & 16.98 & 16.67 & 26.4 & 28.29 & 27.36 & 3.76 & 3.10 & 3.39 & 5.28 & 4.59 & 4.94 \\ 
     & DCRNN & 14.97 & 16.56 & 15.75 & 24.16 & 27.49 & 25.88 & 3.67 & 3.10 & 3.39 & 5.28 & 4.59 & 4.94 \\ 
     & MTGNN & 14.86 & 16.26 & 15.54 & 25.20 & 27.98 & 26.63 & 3.63 & \textbf{3.09} & 3.37 & 5.17 & \textbf{4.52} & 4.85 \\ 
     \midrule
    \multirow{3}{*}{ST Transformer} & GMAN & 14.92 & 16.52 & 15.69 & \textbf{24.56} & 27.60 & 26.14 & 3.59 & 3.10 & 3.35 & 5.18 & 4.65 & 4.92 \\
     & STTN & 15.14 & 16.78 & 15.94 & 24.72 & 27.60 & 26.14 & 3.81 & 3.27 & 3.54 & 5.47 & 4.80 & 5.15 \\ 
     & PDFormer & \textcolor{red}{14.64} & \textbf{16.05} & \textbf{15.33} & \textcolor{red}{23.70} & \textbf{26.61} & \textbf{25.20} & \textbf{3.54} & \textbf{3.07} & \textbf{3.31} & \textbf{5.09} & 4.53 & \textbf{4.81} \\ 
     \midrule
    Ours & STDCformer & \textbf{14.81} & \textcolor{red}{15.67} & \textcolor{red}{15.24} & 24.66 & \textcolor{red}{26.35} & \textcolor{red}{25.17} & \textcolor{red}{3.33} & 3.24 & \textcolor{red}{3.27} & \textcolor{red}{5.08} & \textcolor{red}4.47 & \textcolor{red}{4.77} \\ 
    \bottomrule
\end{tabular}
}
\caption{Multi-step prediction results on MHT and BKL for STDCformer and baselines. Best performance is marked in \textcolor{red}{red}, and the second-best is marked in \textbf{bold}. The metrics on Inflow, Outflow and overall flow are represented by In, Out and IO, respectively.}
\label{tab: tab3}
\end{table}

\subsubsection{Zero-Shot Transfer Experiments Under OOD Condition}
\label{sec: ood}
The evaluation results of the zero-shot experiments under OOD conditions are shown in Table \ref{tab:zero-shot}. STDCformer outperforms the baseline method in all metrics, indicating that STDCformer has better generalization ability. And it can be observed that on the same dataset, the differences in the metrics between models are more significant on the zero-shot task than on the supervised learning task, which implies that the de-confounded model has a greater advantage in generalization ability. The reason behind may be that although the distribution of spatial-temporal confounders in the ood dataset is shifted, frozen confounder encoder is still able to capture the difference of characteristics between STTs and use it to assign weights, which makes the weights in the de-confounding spatial-temporal fusion module is still able to de-confounding and make prediction.

The ability of the model to transfer to an OOD dataset in a zero-shot manner is of significant practical importance, as it allows for introducing models from other regions with minimal cost to support emergent decision-making while achieving acceptable performance. Although the accuracy of this zero-shot testing is higher compared to training from scratch on the OOD dataset, it is still acceptable for predictive needs in emergency situations, meeting the requirements of relevant planning and management departments.

\begin{table}[h]
\centering
\caption{Results of Zero-shot transfer experiments on OOD dataset. Best performance is marked in \textcolor{red}{red}, and the second-best is marked in \textbf{bold}. The result is for overall IO.}
\label{tab:zero-shot}
    \begin{tabular}{ccccc}
    \toprule
    \textbf{Model} & \textbf{MAE} & \textbf{MAPE(\%)} & \textbf{MSE} & \textbf{RMSE} \\ 
    \midrule
    GMAN & 7.69 & 124.41 & 131.05 & 10.97 \\ 
    PDFormer & \textbf{6.94} & \textbf{80.70} & \textbf{95.16} & \textbf{9.23} \\ 
    STDCformer & \textcolor{red}{6.03} & \textcolor{red}{65.41} & \textcolor{red}{95.08} & \textcolor{red}{9.22} \\
    \bottomrule
    \end{tabular}
\end{table}

\subsection{Ablation and Hyperparameters Analysis}

\subsubsection{Ablation Analysis}
\begin{figure}[h]
	\centering
	\subfloat[]{\includegraphics[width=.5\columnwidth]{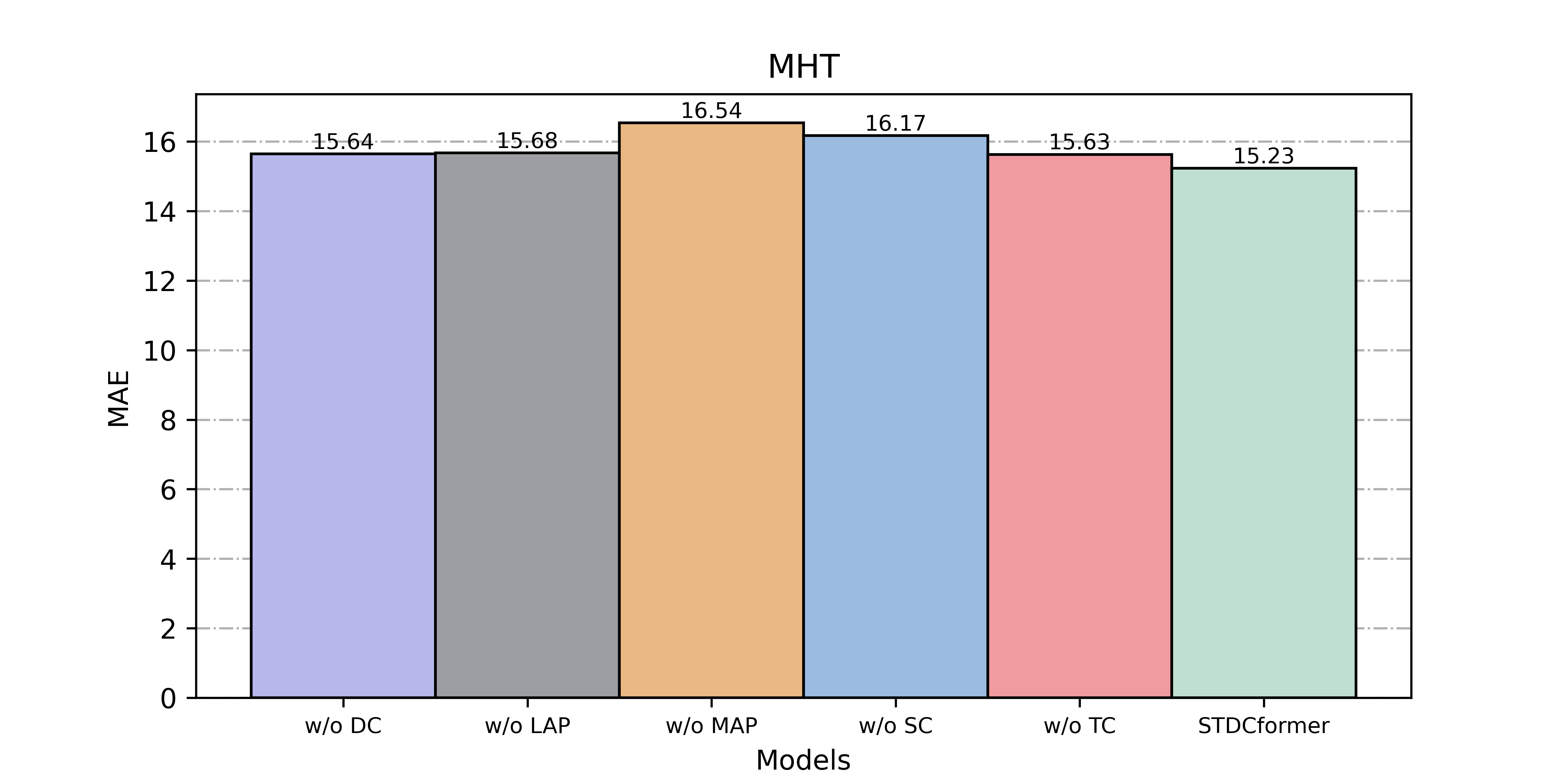}\label{fig: fig12a}}
	\subfloat[]{\includegraphics[width=.5\columnwidth]{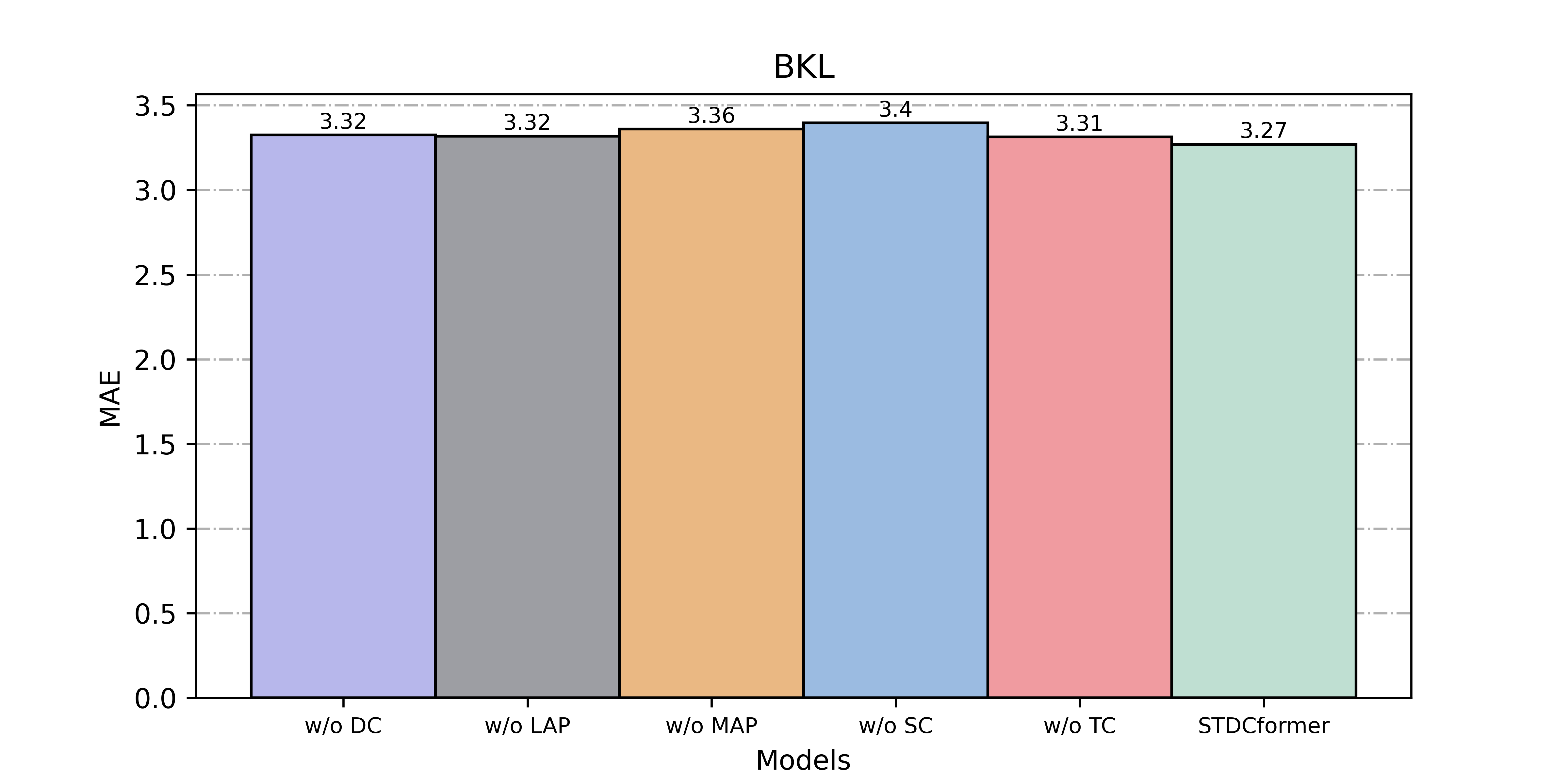}\label{fig: fig12b}}
	\caption{The results of ablation experiments. (a) MHT dataset. (b) BKL dataset.}
    \label{fig: fig12}
\end{figure}

To further validate the effectiveness of the components in STDCformer, we conducted ablation experiments. The MAE of the model after ablation is shown in Figure \ref{fig: fig12}. We set up the following five ablated models:
\begin{enumerate}
    \item 'w/o DC': The original model without the de-confounding module. Ablating this module demonstrates the importance of considering reweighting in the spatial-temporal fusion process.
    \item 'w/o MAP': The original model without CTA-based mapping. Ablating this module highlights the importance of learning the transformation relationship between past and future states in the representation space.
    \item 'w/o SC': The original model without the spatial confounder. Ablating this module shows the significance of integrating auxiliary information to characterize the spatial characteristics of different regions.
    \item 'w/o TC': The original model without the temporal confounder. Ablating this module illustrates the importance of incorporating auxiliary information to represent the temporal characteristics of each time window.
    \item 'w/o LAP': The original model without the Laplacian embedding. Ablating this module emphasizes the importance of retaining the global adjacency graph structure when characterizing spatial properties.
\end{enumerate}

Overall, it can be observed that the model's performance on the BKL dataset does not change significantly, with the original model achieving the best performance. This suggests that the backbone, spatial-temporal transformer, ensures the model's basic capability on simple datasets, and the additional components we designed enhance performance on top of this foundation. In contrast, on the MHT dataset, which presents a greater prediction challenge, the contribution of each component to prediction accuracy provides a clear and intuitive comparison.

After removing the CTA-based mapping module, the model’s accuracy dropped the most, indicating that the differences between the spatial-temporal characteristics of past and future STTs are significant. The mapping relationship is highly complex and diverse, requiring the design of effective modules to uncover these dynamic relationships. The prediction performance also declined after removing the Laplacian embedding and the De-confounding module, demonstrating the importance of incorporating the region’s position within the global spatial structure and accounting for the weights of the temporal and spatial characteristics of the STTs in improving prediction accuracy. The model’s accuracy also showed an evident decline after removing the spatial confounder and temporal confounder, even falling below that of relatively weaker STGNN models. This demonstrates that modeling the auxiliary information as STT characteristics is beneficial for enhancing the model’s predictive capability.

\subsubsection{Hyperparameters Analysis}
We further investigate the sensitivity of the model’s predictions to hyperparameter settings. Since our primary goal is to explore the stability of the model’s performance across different parameter ranges and demonstrate that the model’s effectiveness is not benefit by randomness, we selected hyperparameter ranges that vary around the optimal parameter value. We chose values in a reasonable scope, because extreme values were excluded from consideration by any method.

\begin{figure}[h]
\centering
\includegraphics[width=0.8\textwidth]{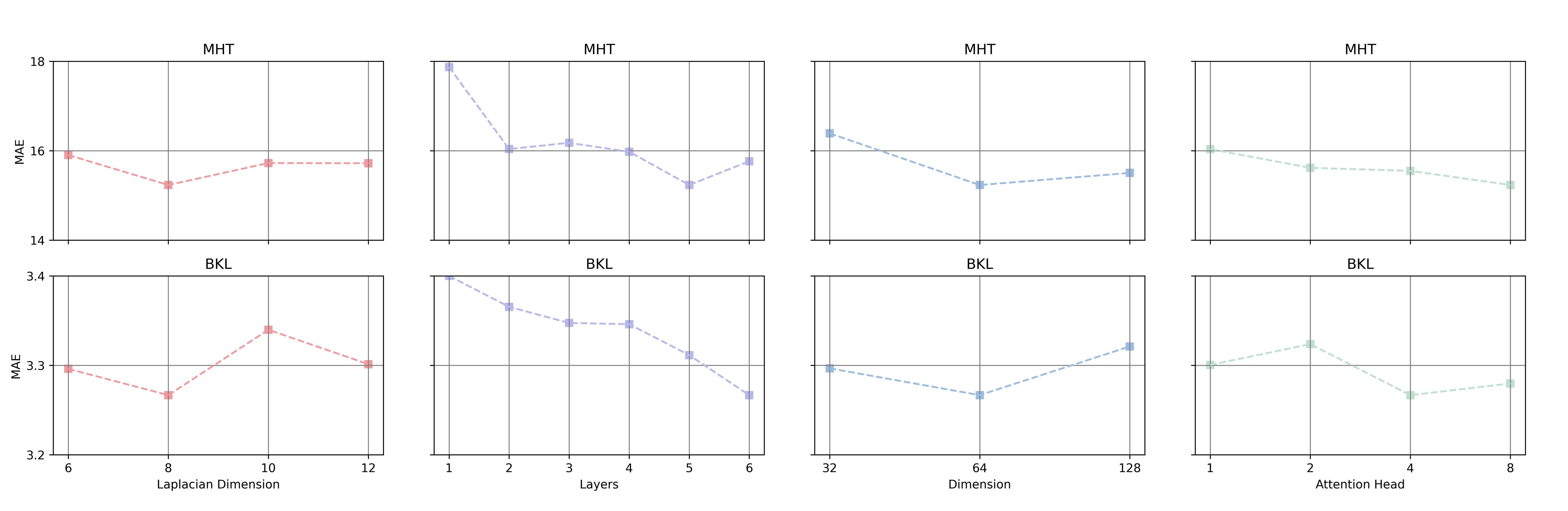}
\caption{The results of hyperparameter analysis.}
\label{fig: fig13}
\end{figure}

The hyperparameters explored include: (1)\textbf{Laplacian Dimension}: exploring the relationship between spatial structural information and model performance through the dimension of Laplacian eigenvalues; (2) \textbf{Layers}: exploring the relationship between model complexity, feature abstraction, and model performance through the number of STDC attention blocks; (3) \textbf{Dimension}: exploring the relationship between the representation space dimension and model performance through the dimension of the hidden representation space; (4) \textbf{Attention Head}: exploring the relationship between the self-attention mechanism and model performance through the number of attention heads. Taking the MAE of the prediction results as an example, the results are shown in Figure \ref{fig: fig13}. Overall, the model exhibits stable performance across different hyperparameter settings, indicating a strong basic predictive ability. The impacts of different hyperparameters are as follows:
\begin{enumerate}
    \item \textbf{The dimension of Laplacian eigenvalues}. Since the Laplacian eigenvalues need to be concatenated with other spatial confounders, we selected values close to and lower than the dimension of the hidden representation space with a scope of [6,8,10,12]. The best performance was achieved when the value was set to 8, suggesting that at this point, the spatial position features and spatial semantic features of different regions might have reached an optimal balance.
    \item \textbf{Number of STDC Attention Blocks}. While increasing the number of layers theoretically enables the model to learn more abstract representations, excessive layers can make the model over-complex, reducing efficiency or even leading to overfitting. Therefore, the maximum number of layers was set to 6. Overall, the predictive ability of the model improved with more layers, with the MHT and BKL datasets achieving optimal performance at 5 and 6 layers, respectively. The optimal values on these two datasets show little difference, however, for more complex datasets with higher prediction difficulty, more layers are required.
    \item \textbf{Dimension of the hidden representation space}. Since the original dimensions of the time-series observations are relatively low, mapping them to excessively high hidden dimensions may lead to information sparsity. Therefore, we chose a range of [32,64,128]. The optimal dimension for the hidden representation space on two datasets was both 64, suggesting that when the original 2-dimensional flow feature is scaled to a 64-dimensional space, an optimal trade-off is achieved between information retention and feature extraction.
    \item \textbf{Number of Attention Heads}. Different attention heads can capture attention coefficients with different semantics. Following existing studies, we set the range to [1,2,4,8]. Multi-head attention outperformed single-head attention, with the MHT and BKL datasets achieving optimal performance at 8 and 4 heads, respectively. This indicates that the semantic complexity of the MHT dataset is higher than that of the BKL dataset, which may be strongly correlated with the prediction difficulty.
\end{enumerate}

\subsection{Analytical Experiments and Interpretation}
To further investigate and explain the reasons behind the model's strong performance, in addition to the zero-shot transfer experiment for spatial OOD regions (Section \ref{sec: ood}), we conducted a series of analytical experiments and provided a physical interpretation aligned with real-world scenes. To validate whether the spatial-temporal de-confounding based fusion approach improves the model's prediction balance and mitigates the impact of data bias, we conducted a statistical analysis and comparison of prediction balance (Section \ref{sec: balance}). To demonstrate the advantages of the proposed model in real-world predictions, we performed an analysis and interpretation based on prediction visualization (Section \ref{sec: visualization}). To further explore the significance of the two key issues in spatial-temporal prediction proposed in Section \ref{sec: hypothesis1} and Section \ref{sec: hypothesis2} in real-world tasks, as well as their physical meaning, we also conducted a case study of the research region (Section \ref{sec: case1} and Section \ref{sec: case2}).

\subsubsection{Prediction Balance}
\label{sec: balance}
\begin{figure}[h]
	\centering
	\subfloat[]{\includegraphics[width=.5\columnwidth]{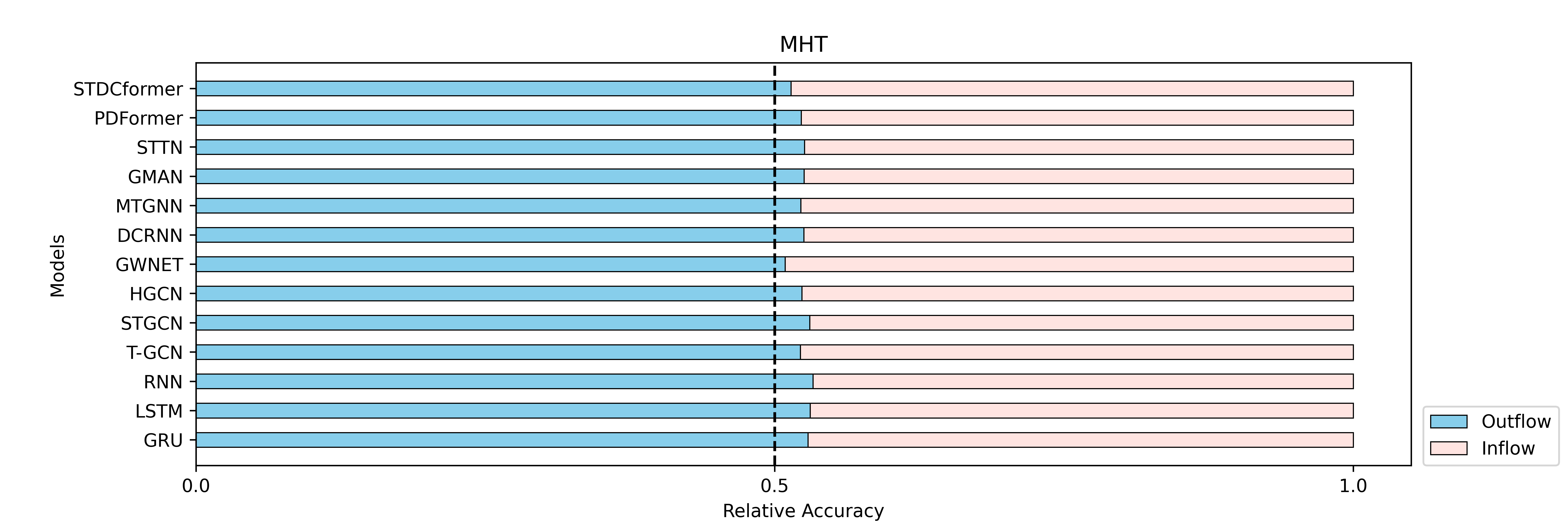}\label{fig: fig14a}}
	\subfloat[]{\includegraphics[width=.5\columnwidth]{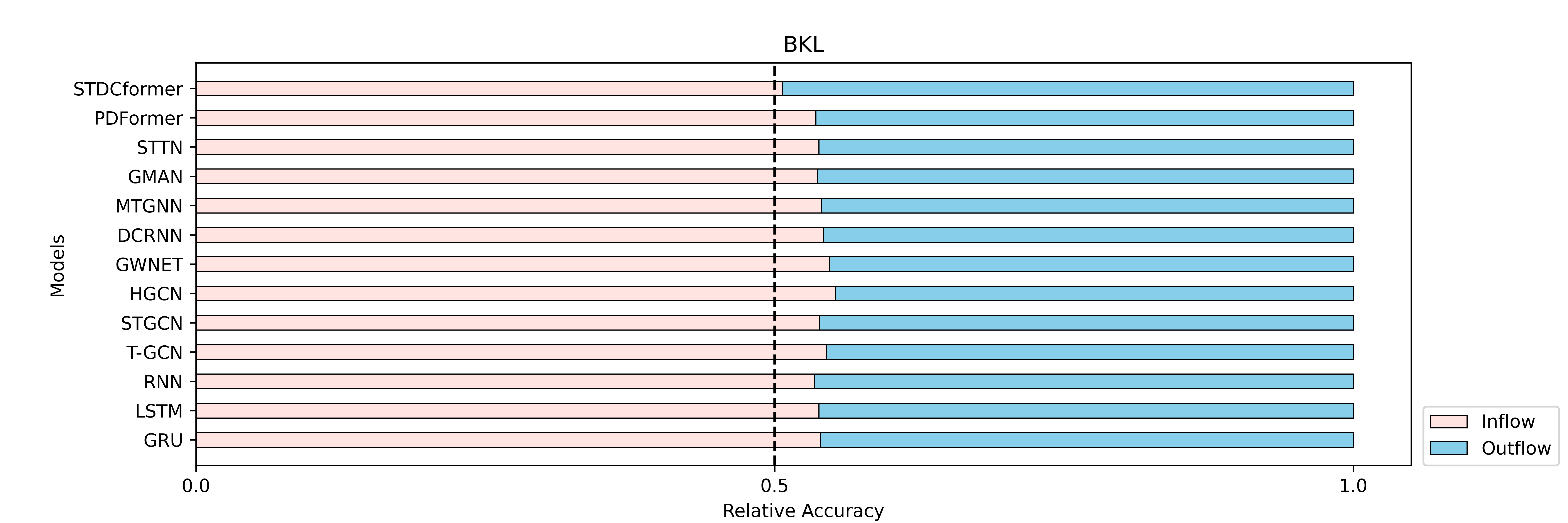}\label{fig: fig14b}}
	\caption{The comparison of prediction accuracy of Inflow and Outflow. (a) MHT dataset. (b) BKL dataset.}
    \label{fig: fig14}
\end{figure}

\textbf{Balance between Inflow and outflow}. Crowd flow prediction requires capturing both inflow and outflow characteristics of a region, making it a multi-feature prediction task. Different models perform differently on inflow and outflow predictions and often fail to achieve a balanced performance. This imbalance arises because inflow and outflow can exhibit distinct patterns for the same region. For instance, during working hours, a company may load higher inflow than outflow, while the opposite may occur during after-work hours. Models’ predictive imbalance across features may be raised by biases in the distribution of features within the dataset, causing the model’s favor of certain features. Figure \ref{fig: fig14} illustrates the relative MAE ratio of predicting inflow and outflow by different models. We observe that STDCformer achieves the most balanced performance between inflow and outflow predictions across both datasets. This balance may be attributed to the incorporation of spatial characteristics, which help in constructing a semantic profile of the region (e.g., functionality). These semantics act as common driving factors behind inflow and outflow variations, enabling the model to capture changes in these features more effectively.

\begin{figure}[h]
\centering
\includegraphics[width=0.8\textwidth]{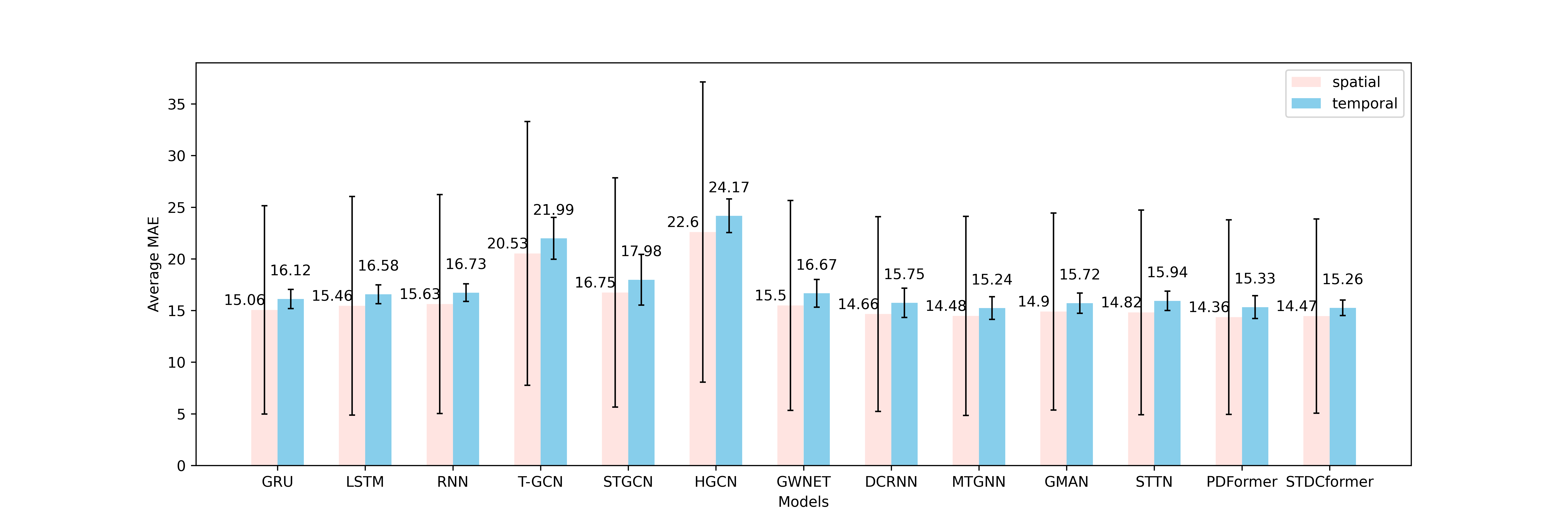}
\caption{Models’ prediction accuracy across different time windows and regions in MHT dataset.}
\label{fig: fig15}
\end{figure}

\textbf{Balance across regions and across timesteps}. As shown in Figure \ref{fig: fig15}, using MHT dataset as an example, we examine the distribution of the model's performance across spatial and temporal dimensions for each model. We gathered the statistic of prediction metrics separately for different regions and time windows, and report the mean and standard deviation. Taking the difference in MAE as an example, it can be observed that STDCformer achieves the smallest standard deviation in prediction across both regions and time steps. This indicates that STDCformer is able to maintain the best balance across spatial and temporal dimensions. The underlying reason for this performance is likely the incorporation of spatial-temporal confounders’ information, followed by a de-confounding process. This approach allows the model to distinguish samples from different spatial and temporal contexts, rather than treating each spatial-temporal sample equally in the parameter fitting process.

\subsubsection{Prediction Visualization}
\label{sec: visualization}
\begin{figure}[h]
\centering
\includegraphics[width=0.8\textwidth]{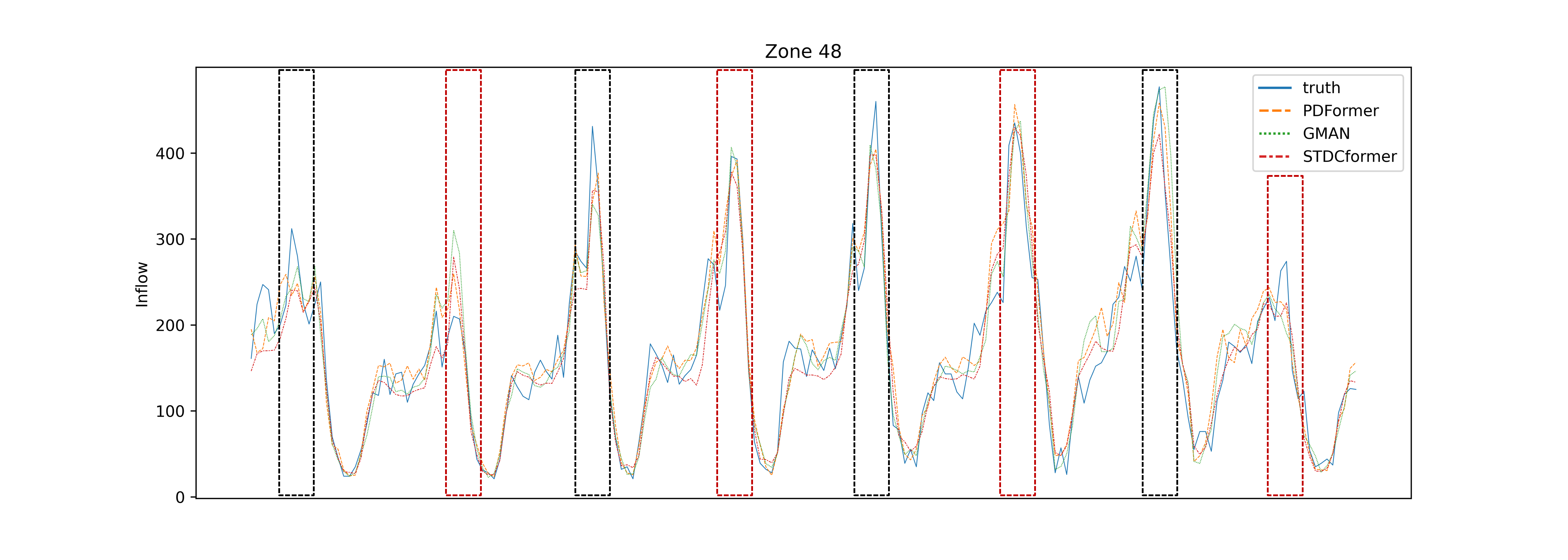}
\caption{ Visualization of prediction result for Zone 48.}
\label{fig: fig16}
\end{figure}

\begin{figure}[h]
\centering
\includegraphics[width=0.8\textwidth]{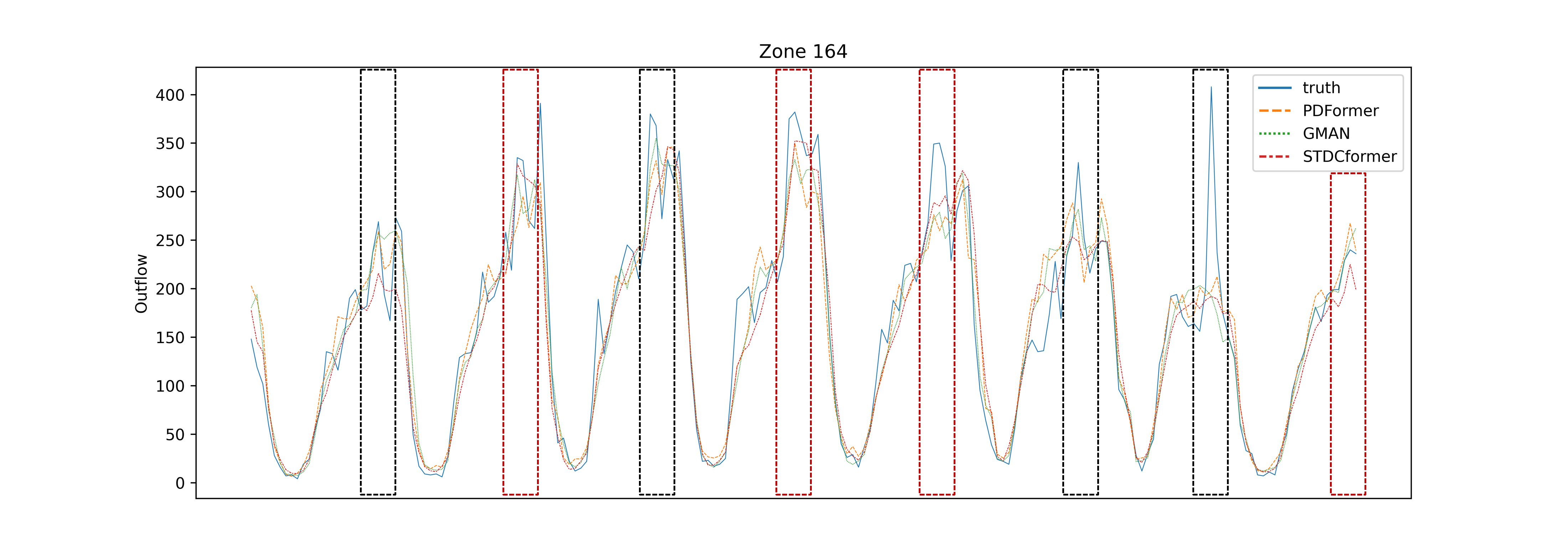}
\caption{Visualization of prediction result for Zone164.}
\label{fig: fig17}
\end{figure}

In general, all models perform well in capturing the upward and downward trends of crowd flow. However, STDCformer prefer to make predicted values within a relatively small range, likely due to the inherent restraint between the spatial confounder $C_S$ and temporal confounder $C_T$. Because STDCformer places significant importance on spatial characteristics, which are considered stable, thus leading to more conservative predictions. For traffic flow peaks that last for over two hours (as indicated by the red dashed boxes in Figures \ref{fig: fig16} and Figures \ref{fig: fig17}), STDCformer provides relatively accurate predictions. On the other hand, PDFormer and GMAN tend to predict values that exceed the actual peak. However, in the case of sudden increases followed by sharp decreases in crowd flow (as indicated by the black dashed boxes in Figures \ref{fig: fig16} and Figures \ref{fig: fig17}), STDCformer struggles to make more accurate predictions for the peak. This observation indicates two points, on the one hand, STDCformer is less sensitive to temporal changes, and its ability to predict flow shifts caused by unexpected events is relatively weaker. On the other hand, when the traffic flow data contains errors or noise (such as incorrect passenger counts for taxi orders), STDCformer is better at filtering out the noise.

\subsubsection{Physical Meaning of Confounders}
\label{sec: case1}
To explore the physical meaning of the temporal and spatial confounders learned by the model, we analyzed the weights output from the spatial-temporal de-confounding layers. Here, $P\left(C_S\right)$ represents the weight of the spatial confounder in the STTs. Based on the observed phenomena, we hypothesize and provide an interpretation of the role and potential implications of the learned confounders. From the perspective of de-confounding, higher weights indicate that these confounders represent a lower proportion of the original data. Therefore, higher weights are needed to model an ideal unbiased distribution. From the perspective of spatial-temporal data fusion, higher weights suggest that, under the original data distribution, the model needs to pay more attention to the information embedded within this type of confounders.

\begin{enumerate}
    \item 
    \textbf{Observation 1}: For all regions, $P\left(C_S\right)$ in the training dataset is no less than 0.5.

    \textbf{Guess}: The original data distribution leads the model to be more biased toward capturing temporal representations without a de-confounding strategy. Introducing higher weight of $P\left(C_S\right)$ allows the model to emphasize the role of spatial representations during the spatial-temporal fusion process.

    \textbf{Interpretation}: When predicting future crowd flow, the model considers the differences between regions and utilizes the position of each region within the global urban context for prediction. The learned ${P}\left({C}_{S}\right)$ in this work not only captures the spatial characteristics of regions but also includes the inter-region connectivity, which preserves the structural relationships between spatial regions from both an attribute and a positional perspective. On one hand, ${P}\left({C}_{S}\right)\geq0.5$ highlights the importance of spatial information in spatial-temporal predictions, which is consistent with previous observations, such as the pre-experiment results from STGC-GNNs\cite{RN50}, which demonstrated the importance of spatial information, and the ablation study results from PDFormer\cite{RN18}, which also emphasizes this. On the other hand, ${P}\left({C}_{S}\right)\geq0.5$ also implies that ${P}\left({C}_{S}\right)\geq{P}\left({C}_{T}\right)$, indicating that the temporal confounder has a higher weight in the original data distribution. This leads the model without de-confounding to rely more heavily on temporal features. The de-confounding process in this work adjusts this bias and corrects the overemphasis on temporal representations.

    \item 
    \textbf{Observation 2}: When the IO value decreases, ${P}\left({C}_{S}\right)$ increases (as shown as Figure \ref{fig: fig18}).

    \textbf{Guess}: The enhancement of ${C}_{S}$ can serve as an effective buffer when the functional characteristics of a region diminish.

    \textbf{Interpretation}: For a spatial region, a decrease in crowd flow suggests that its functionality is less being utilized, causing the region’s spatial portray increasingly vague. The differences between regions, as modeled by spatial confounders, are reduced. This leads the model to rely more heavily on temporal confounders, further amplifying the bias in the distribution of spatial and temporal confounders in the data. By explicitly incorporating spatial features as confounders, this work ensures that the spatial characteristics of each region remain relatively stable. Thus, when the weight of a region's functionality in the observational data decreases, ${C}_{S}$ can counterbalance this by leveraging its interplay with ${C}_{T}$, preserving the utilization of spatial characteristics. Strengthening the use of spatial confounder’s information helps to retain the spatial information and prevent a decline in predictive accuracy.

    \begin{figure}[h]
    	\centering
    	\subfloat[]{\includegraphics[width=.33\columnwidth]{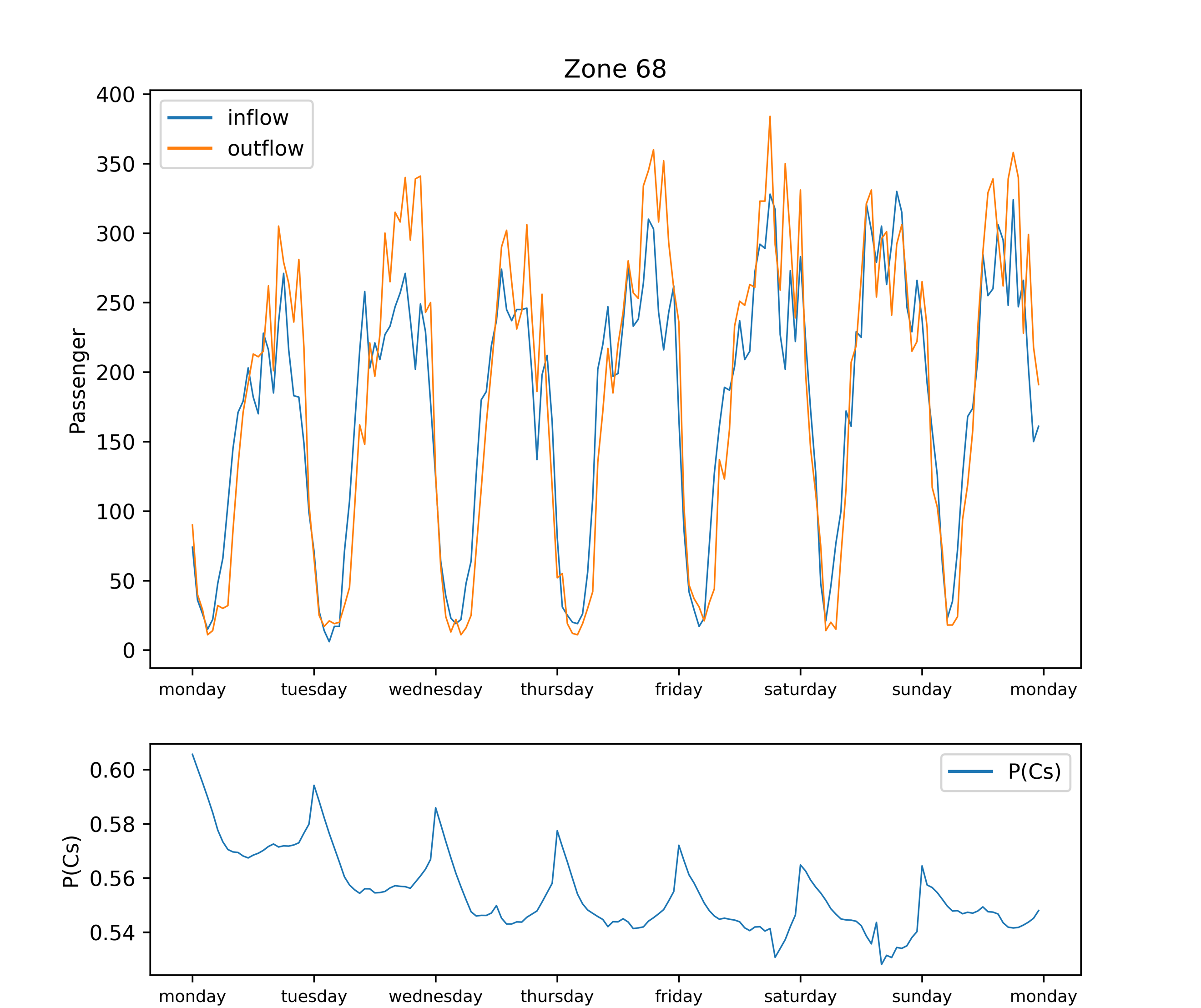}\label{fig: fig18a}}
    	\subfloat[]{\includegraphics[width=.33\columnwidth]{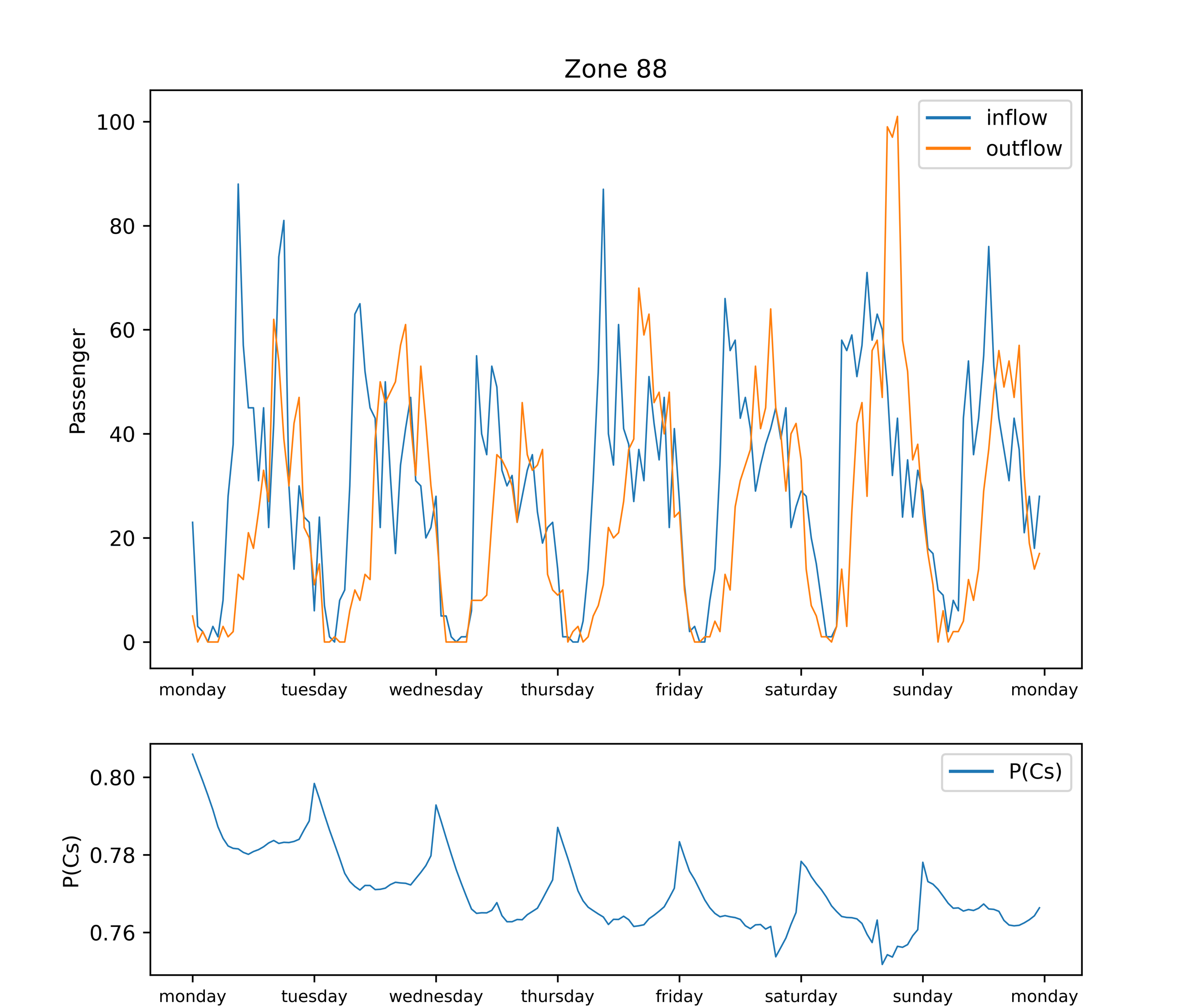}\label{fig: fig18b}}
        \subfloat[]{\includegraphics[width=.33\columnwidth]{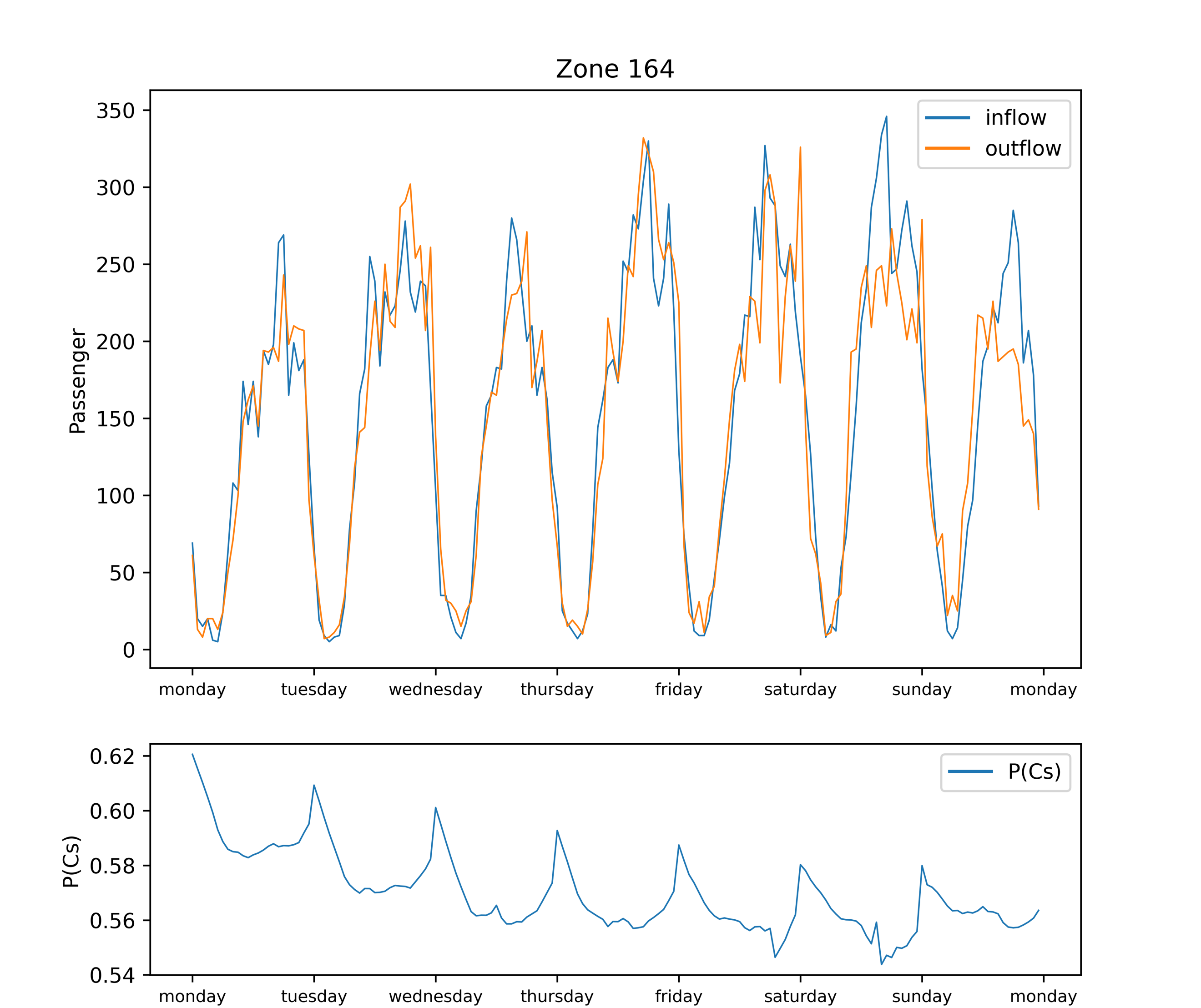}\label{fig: fig18c}}
    	\caption{Trend variation of crowd flow and ${P}\left({C}_{S}\right)$ over time. (a) Zone 68. (b) Zone 88. (c) Zone 164}
        \label{fig: fig18}
    \end{figure}

    \item 
    \textbf{Observation 3}: For regions in closer clusters, the values of their ${P}\left({C}_{S}\right)$ are also closer.

    \textbf{Guess}: The weight of ${C}_{S}$ for a region is significantly associated with its spatial characteristics and may serve as an indicator of the region's functionality.

    \textbf{Interpretation}: As shown in Figure \ref{fig: fig19} and Figure \ref{fig: fig20}, regions are clustered into six categories based on their POI features. It can be observed that regions with similar clustering results also exhibit closer average $P\left(C_S\right)$ values. The most likely reason is that these regions share similar functionality. To further explore this relationship, we selected several representative regions for detailed analysis.
    \begin{itemize}
        \item Roosevelt Island (Zone 194) and Randall’s Island (Zone 202). Both regions exhibit high $P\left(C_S\right)$ values and share similar functional zoning. Visualization of crowd flow data reveals that these regions are visited infrequently and often experience periods with no arrivals or departures. This is due to their highly singular and specialized functions, primarily providing recreational and leisure spaces. The confounding bias in the original data tends to downplay the importance of spatial characteristics for these regions. As a result, the model compensates by enhancing the spatial features for such areas, assigning them higher $P\left(C_S\right)$. Similarly, Highbridge Park (Zone 120) and The Battery Park (Zone 12 \& Zone 13) also exhibit high $P\left(C_S\right)$ values due to their analogous characteristics. In contrast, Central Park (Zone 43), although functionally similar to these regions, has a relatively low $P\left(C_S\right)$ value. This can be attributed to its location near the center of Manhattan, which makes it highly accessible and surrounded by high human flow. As a result, the dependence of crowd flow on spatial functionality is reduced for this area.
        
        \item Midtown (e.g., Zone 164, Zone 48, Zone 68). The zones in midtown exhibit generally smaller $P\left(C_S\right)$, with values in the weekends lower than those on weekdays. This is due to the rich variety of functionalities in these areas, including numerous tourist attractions, department stores, and restaurants, which attract a large number of visitors. Consequently, the overall crowd flow is very high, and there is little difference between weekdays and weekends. For such regions, the overall spatial functionality is accessed relatively consistently, meaning there is less need to enhance the utilization of spatial characteristics. Instead, the model relies more on temporal trends for accurate predictions. A similar phenomenon can also be observed within the downtown area.
    \end{itemize}
\end{enumerate}

\begin{figure}[h]
	\centering
	\begin{minipage}[c]{0.4\textwidth}
		\centering
		\includegraphics[height=0.8\textwidth]{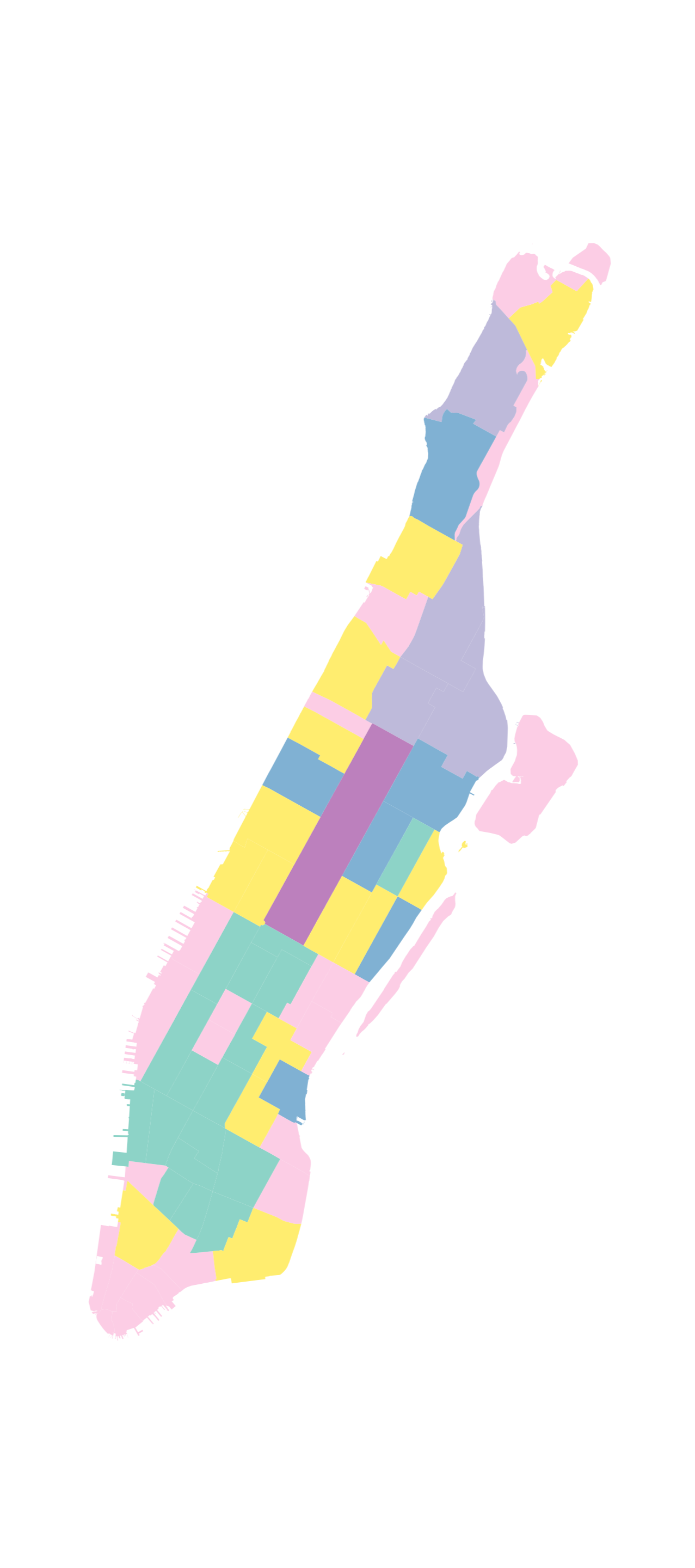}
		\caption{Clusters distribution of MHT based on POI.}
		\label{fig: fig19}
	\end{minipage} 
	\begin{minipage}[c]{0.4\textwidth}
		\centering
		\includegraphics[height=0.8\textwidth]{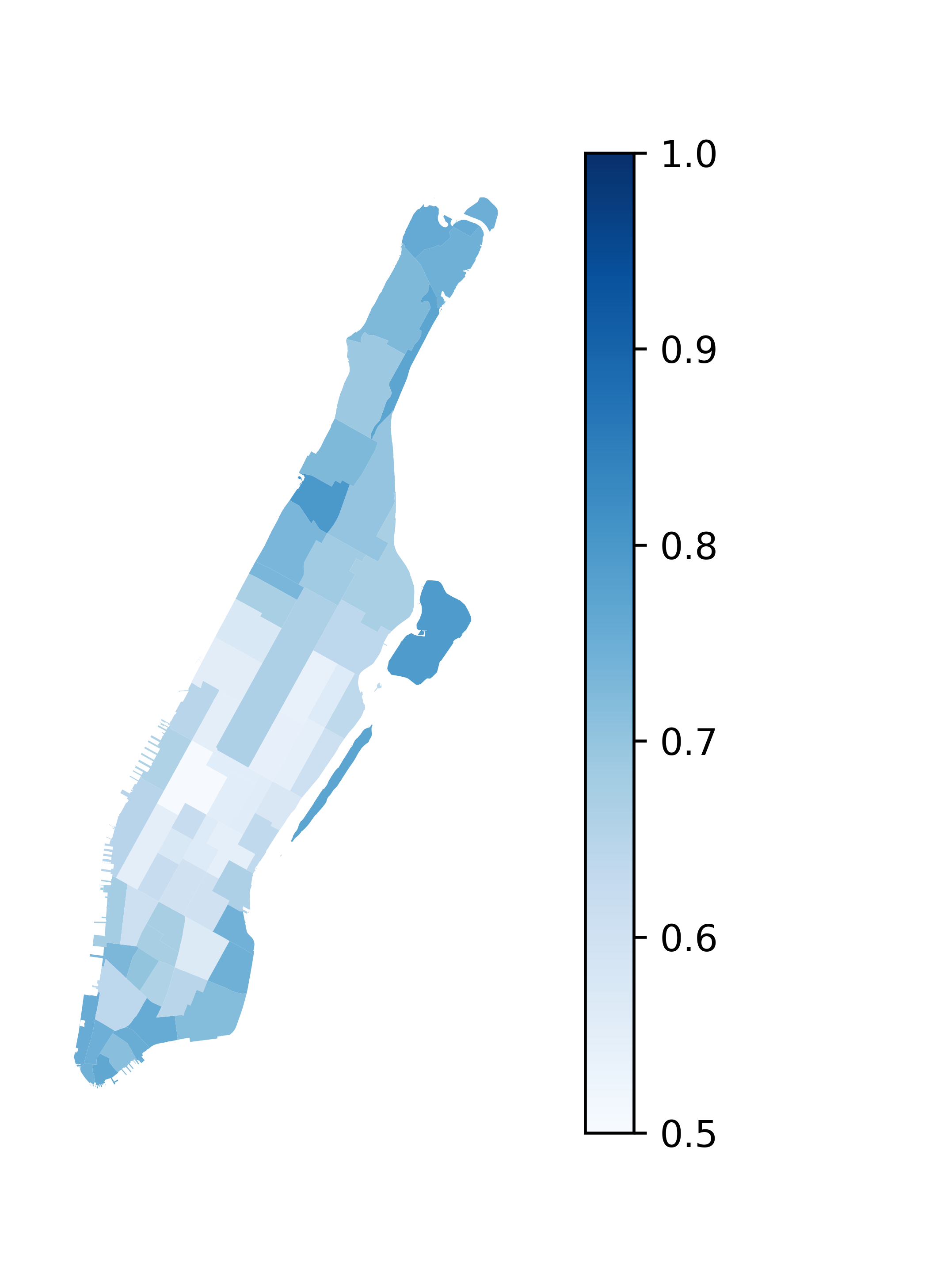}
		\caption{The distribution of average value of $P\left(C_S\right)$ in MHT.}
		\label{fig: fig20}
	\end{minipage}
\end{figure}



\subsubsection{Physical Meaning of Cross-Time Attention}
\label{sec: case2}
\begin{figure}[h]
	\centering
	\subfloat[]{\includegraphics[width=.33\columnwidth]{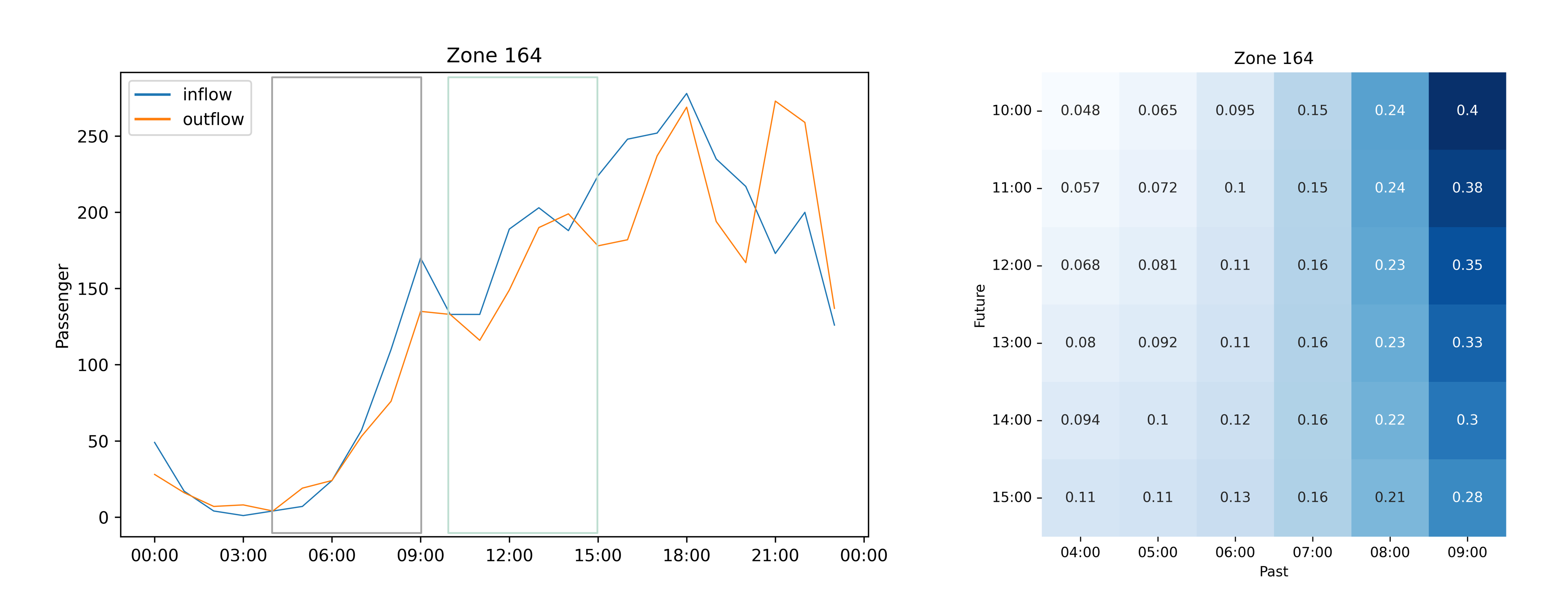}\label{fig: fig21a}}
	\subfloat[]{\includegraphics[width=.33\columnwidth]{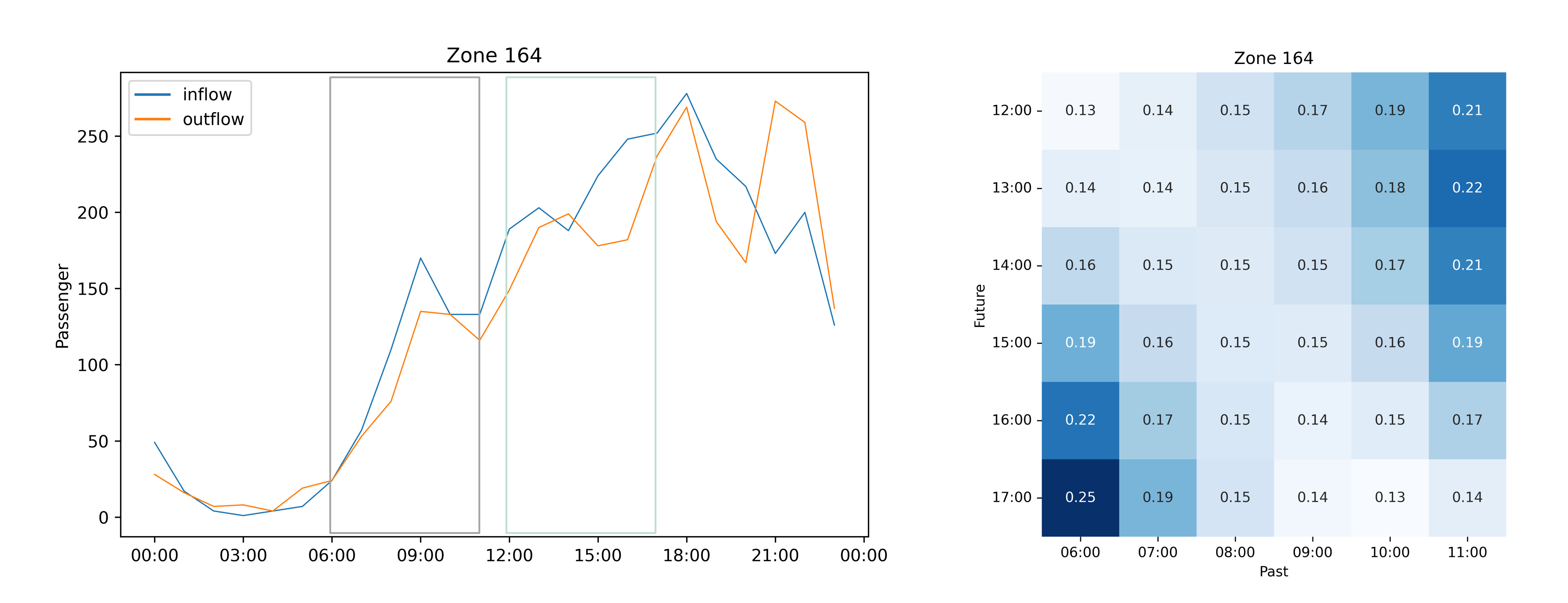}\label{fig: fig21b}}
    \subfloat[]{\includegraphics[width=.33\columnwidth]{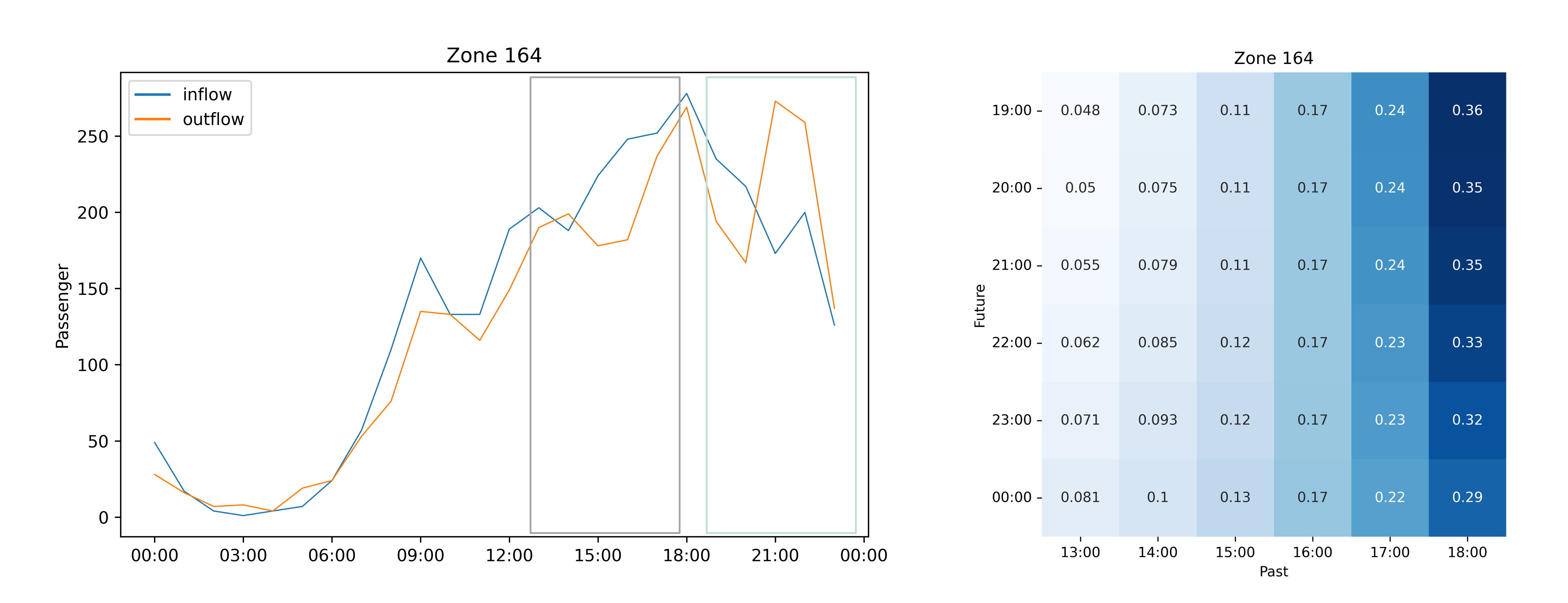}\label{fig: fig21c}}
	\caption{The flow variation of Zone164 and corresponding cross-time attention.}
    \label{fig: fig21}
\end{figure}

Once the window sizes for past and future time intervals are determined, the prediction task can be viewed as two sliding windows (past, value) along the time axis. Different windows can capture varying influences of the past on the future. For the same spatial region, the dependency between the future and the past also differs across time periods. Taking Zone 164 as an example, the cross-time mapping attention under different (past, future) windows is shown in Figure \ref{fig: fig21}, where the gray box represents the past time window, and the green box represents the future time window. Since crowd flow in more recent time steps may be more similar, future time steps are likely to be more dependent on the last time step in the past window, which leads to assigning more attention to later time steps. When the future trend is very similar to the historical trend (Figure \ref{fig: fig21a}) or completely dissimilar ((Figure \ref{fig: fig21b}), the model tends to assign more attention to the most recent historical data. However, as shown in Figure \ref{fig: fig21c}, as the prediction horizon increases, the similarity between the latter part of the future data and the historical data decreases. The model starts to focus more on the similarity of trends between past and future. Therefore, when the latter part of the future data shows a similar trend to some historical data segment, the model may capture this similarity and assign more attention to more distant historical data during prediction.

\begin{figure}[h]
\centering
\includegraphics[width=0.6\textwidth]{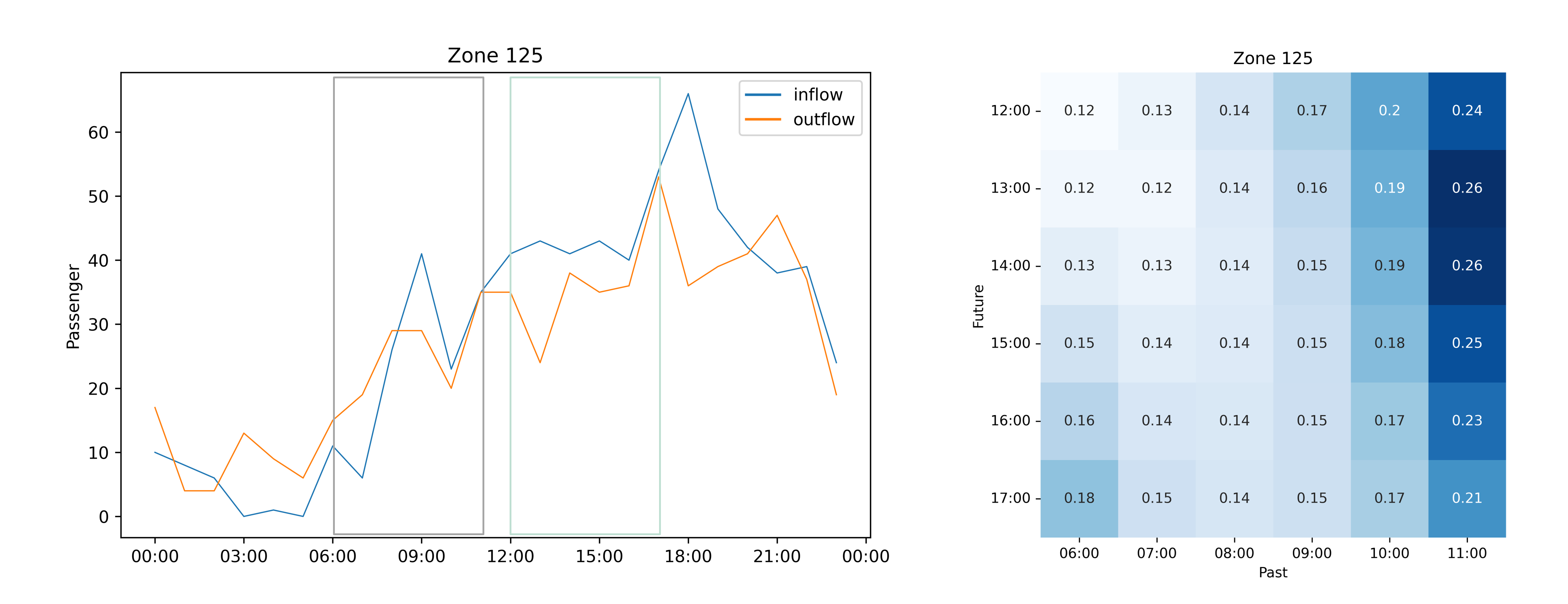}
\caption{The flow variation of Zone 125 and corresponding cross-time attention.}
\label{fig: fig22}
\end{figure}

For different spatial regions during the same time period, the influence of the past on the future also varies. As shown in Figure \ref{fig: fig22}, the values in the future window for Zone 125 exhibit relatively stable changes, remaining close to the value of the last historical time step. Consequently, the future time steps are more reliant on the values from the most recent historical time steps. While there is a similar trend of attention shifting toward more distant historical time steps, as observed earlier, the degree of this shift is less pronounced. These observations align with our hypothesis that the mapping relationships between the past and the future vary across different STTs.

\section{Conclusion and Future Works}
Crowd Flow Prediction, as a typical spatial-temporal prediction task, holds significant research and application value. We decompose the crowd flow prediction task into two key components: the space transformations between the observation space and the representation space, and the spatial-temporal mapping from the past to the future. Correspondingly, two important hypotheses are proposed: (1) learning a Spatial-Temporal De-Confounded (STDC) representation space, and (2) establishing a Past-to-Future mapping based on a cross-time attention mechanism. To learn the de-confounded representation space, we introduce a new Spatial-Temporal Backdoor Adjustment Strategy, which divides hidden confounders into temporal confounders and spatial confounders. This strategy can control confounding effects during the spatial-temporal fusion process. To learn the Past-to-Future Mapping, we construct a Spatial-Temporal Embedding based on spatial-temporal confounders to encode the spatial-temporal characteristics of the past and the future representations. A Cross-Time Attention mechanism is then proposed to query the relationship between the past and future. To better integrate de-confounding with spatial-temporal prediction, we build a spatial-temporal prediction model, STDCformer, using a Transformer architecture as the backbone. This model uses the information of confounders to simultaneously guide spatial-temporal representation learning and spatial-temporal mapping capture, achieving end-to-end de-confounded spatial-temporal prediction with high information efficiency. Experiments on two real-world datasets demonstrate that the proposed STDCformer achieves SOTA performance in crowd flow prediction. Additionally, it shows stronger zero-shot generalization ability on spatial OOD datasets. A series of analytical and visualization experiments further explain the model's predictive performance. We have constructed two real-world crowd flow datasets and provided corresponding auxiliary data to facilitate learning the representations of confounders, which can serve as a valuable resource for related research.

However, crowd flow prediction and other spatial-temporal prediction tasks involve modeling and understanding complex spatial-temporal processes, particularly in urban scenes where man-earth interactions further complicate the problem. A key challenge in these tasks is how to characterize the numerous influence factors underlying spatial-temporal processes and, based on this, understand the dynamic and complex interactions between human behavior and geographical locations. This presents a critical breakthrough point for such prediction tasks. In the future, we plan to leverage the powerful reasoning capabilities of large language models (LLMs) to infer the underlying influence factors from text. Additionally, we aim to utilize the strong representation capabilities of LLMs to obtain richer representations of confounders, which can be ensured through multimodal alignment strategies.





\section*{Acknowledgement}
This research was funded by the National Natural Science Foundation of China under Grant 42271481 and the Natural Science Foundation of Hunan Province under Grant 2022JJ30698. This work was carried out in part using computing resources at the High Performance Computing Platform of Central South University.

\bibliographystyle{elsarticle-num}
\bibliography{references}

\begin{thebibliography}{10}
\expandafter\ifx\csname url\endcsname\relax
  \def\url#1{\texttt{#1}}\fi
\expandafter\ifx\csname urlprefix\endcsname\relax\def\urlprefix{URL }\fi
\expandafter\ifx\csname href\endcsname\relax
  \def\href#1#2{#2} \def\path#1{#1}\fi

\bibitem{RN1}
H.~Go, S.~Park, \href{https://doi.org/10.1038/s41598-024-63310-6}{A study on deep learning model based on global–local structure for crowd flow prediction}, Scientific Reports 14~(1) (2024) 12623.
\newblock \href {http://dx.doi.org/10.1038/s41598-024-63310-6} {\path{doi:10.1038/s41598-024-63310-6}}.
\newline\urlprefix\url{https://doi.org/10.1038/s41598-024-63310-6}

\bibitem{RN2}
P.~Xie, T.~Li, J.~Liu, S.~Du, X.~Yang, J.~Zhang, Urban flow prediction from spatiotemporal data using machine learning: A survey, Information Fusion 59 (2020) 1--12.

\bibitem{RN3}
J.~Wang, J.~Jiang, W.~Jiang, C.~Li, W.~X. Zhao, Libcity: An open library for traffic prediction (2021).
\newblock \href {http://dx.doi.org/10.1145/3474717.3483923} {\path{doi:10.1145/3474717.3483923}}.

\bibitem{RN4}
G.~Jin, Y.~Liang, Y.~Fang, Z.~Shao, J.~Huang, J.~Zhang, Y.~Zheng, Spatio-temporal graph neural networks for predictive learning in urban computing: A survey, IEEE Transactions on Knowledge and Data Engineering 36~(10) (2024) 5388--5408.
\newblock \href {http://dx.doi.org/10.1109/TKDE.2023.3333824} {\path{doi:10.1109/TKDE.2023.3333824}}.

\bibitem{RN5}
H.~Liu, C.~Zhu, D.~Zhang, Q.~Li, Attention-based spatial-temporal graph convolutional recurrent networks for traffic forecasting, in: International Conference on Advanced Data Mining and Applications, Springer, 2023, pp. 630--645.

\bibitem{RN6}
Y.~Li, R.~Yu, C.~Shahabi, Y.~Liu, Diffusion convolutional recurrent neural network: Data-driven traffic forecasting (2018).

\bibitem{RN7}
J.~Zhu, X.~Han, H.~Deng, C.~Tao, L.~Zhao, P.~Wang, T.~Lin, H.~Li, \href{https://doi.org/10.1109/TITS.2021.3136287}{Kst-gcn: A knowledge-driven spatial-temporal graph convolutional network for traffic forecasting}, IEEE Transactions on Intelligent Transportation Systems 23~(9 
\newblock \href {http://dx.doi.org/10.1109/tits.2021.3136287} {\path{doi:10.1109/tits.2021.3136287}}.
\newline\urlprefix\url{https://doi.org/10.1109/TITS.2021.3136287}

\bibitem{RN8}
H.~Yan, X.~Ma, Z.~Pu, Learning dynamic and hierarchical traffic spatiotemporal features with transformer, IEEE Transactions on Intelligent Transportation Systems 23~(11) (2022) 22386--22399.
\newblock \href {http://dx.doi.org/10.1109/TITS.2021.3102983} {\path{doi:10.1109/TITS.2021.3102983}}.

\bibitem{RN9}
X.~Yin, G.~Wu, J.~Wei, Y.~Shen, H.~Qi, B.~Yin, \href{https://www.sciencedirect.com/science/article/pii/S0925231220318312}{Multi-stage attention spatial-temporal graph networks for traffic prediction}, Neurocomputing 428 (2021) 42--53.
\newblock \href {http://dx.doi.org/https://doi.org/10.1016/j.neucom.2020.11.038} {\path{doi:https://doi.org/10.1016/j.neucom.2020.11.038}}.
\newline\urlprefix\url{https://www.sciencedirect.com/science/article/pii/S0925231220318312}

\bibitem{RN10}
L.~Zhao, Y.~Song, C.~Zhang, Y.~Liu, P.~Wang, T.~Lin, M.~Deng, H.~Li, T-gcn: A temporal graph convolutional network for traffic prediction, IEEE Transactions on Intelligent Transportation Systems 21~(9) (2019) 3848--3858.

\bibitem{RN11}
Z.~Pan, Y.~Liang, W.~Wang, Y.~Yu, Y.~Zheng, J.~Zhang, Urban traffic prediction from spatio-temporal data using deep meta learning, in: Proceedings of the 25th ACM SIGKDD international conference on knowledge discovery and data mining, 2019, pp. 1720--1730.

\bibitem{RN12}
S.~Guo, Y.~Lin, N.~Feng, C.~Song, H.~Wan, Attention based spatial-temporal graph convolutional networks for traffic flow forecasting, in: Proceedings of the AAAI conference on artificial intelligence, Vol.~33, 2019, pp. 922--929.

\bibitem{RN13}
Z.~Wu, S.~Pan, G.~Long, J.~Jiang, C.~Zhang, Graph wavenet for deep spatial-temporal graph modeling (2019).

\bibitem{RN14}
K.~Guo, Y.~Hu, Y.~Sun, S.~Qian, J.~Gao, B.~Yin, Hierarchical graph convolution network for traffic forecasting, in: Proceedings of the AAAI conference on artificial intelligence, Vol.~35, 2021, pp. 151--159.

\bibitem{RN15}
M.~Li, Z.~Zhu, \href{https://ojs.aaai.org/index.php/AAAI/article/view/16542}{Spatial-temporal fusion graph neural networks for traffic flow forecasting}, Proceedings of the AAAI Conference on Artificial Intelligence 35~(5) (2021) 4189--4196.
\newline\urlprefix\url{https://ojs.aaai.org/index.php/AAAI/article/view/16542}

\bibitem{RN16}
B.~Yu, H.~Yin, Z.~Zhu, Spatio-temporal graph convolutional networks: a deep learning framework for traffic forecasting (2018).

\bibitem{RN17}
M.~Xu, W.~Dai, C.~Liu, X.~Gao, W.~Lin, G.-J. Qi, H.~Xiong, Spatial-temporal transformer networks for traffic flow forecasting, arXiv preprint arXiv:2001.02908.

\bibitem{RN18}
J.~Jiang, C.~Han, W.~X. Zhao, J.~Wang, Pdformer: Propagation delay-aware dynamic long-range transformer for traffic flow prediction, in: AAAI, AAAI Press, 2023.

\bibitem{RN19}
Q.~Luo, S.~He, X.~Han, Y.~Wang, H.~Li, \href{https://www.sciencedirect.com/science/article/pii/S0950705124002727}{Lsttn: A long-short term transformer-based spatiotemporal neural network for traffic flow forecasting}, Knowledge-Based Systems 293 (2024) 111637.
\newblock \href {http://dx.doi.org/https://doi.org/10.1016/j.knosys.2024.111637} {\path{doi:https://doi.org/10.1016/j.knosys.2024.111637}}.
\newline\urlprefix\url{https://www.sciencedirect.com/science/article/pii/S0950705124002727}

\bibitem{RN20}
B.~Huang, H.~Dou, Y.~Luo, J.~Li, J.~Wang, T.~Zhou, Adaptive spatiotemporal transformer graph network for traffic flow forecasting by iot loop detectors, IEEE Internet of Things Journal 10~(2) (2022) 1642--1653.

\bibitem{RN21}
J.~Ma, R.~Guo, C.~Chen, A.~Zhang, J.~Li, \href{https://doi.org/10.1145/3437963.3441818}{Deconfounding with networked observational data in a dynamic environment} (2021).
\newblock \href {http://dx.doi.org/10.1145/3437963.3441818} {\path{doi:10.1145/3437963.3441818}}.
\newline\urlprefix\url{https://doi.org/10.1145/3437963.3441818}

\bibitem{RN22}
Y.~Xia, Y.~Liang, H.~Wen, X.~Liu, K.~Wang, Z.~Zhou, R.~Zimmermann, Deciphering spatio-temporal graph forecasting: A causal lens and treatment, Advances in Neural Information Processing Systems 36.

\bibitem{RN23}
J.~Ji, W.~Zhang, J.~Wang, Y.~He, C.~Huang, Self-supervised deconfounding against spatio-temporal shifts: Theory and modeling, arXiv preprint arXiv:2311.12472.

\bibitem{RN24}
X.~Luo, C.~Zhu, D.~Zhang, Q.~Li, Stg4traffic: A survey and benchmark of spatial-temporal graph neural networks for traffic prediction, arXiv preprint arXiv:2307.00495.

\bibitem{RN25}
S.~Rahmani, A.~Baghbani, N.~Bouguila, Z.~Patterson, Graph neural networks for intelligent transportation systems: A survey, IEEE Transactions on Intelligent Transportation Systems 24~(8) (2023) 8846--8885.
\newblock \href {http://dx.doi.org/10.1109/TITS.2023.3257759} {\path{doi:10.1109/TITS.2023.3257759}}.

\bibitem{RN26}
B.~Gu, J.~Zhan, S.~Gong, W.~Liu, Z.~Su, M.~Guizani, A spatial-temporal transformer network for city-level cellular traffic analysis and prediction, IEEE Transactions on Wireless Communications 22~(12) (2023) 9412--9423.
\newblock \href {http://dx.doi.org/10.1109/TWC.2023.3270441} {\path{doi:10.1109/TWC.2023.3270441}}.

\bibitem{RN27}
A.~Liu, Y.~Zhang, \href{https://www.sciencedirect.com/science/article/pii/S0957417424007504}{An efficient spatial-temporal transformer with temporal aggregation and spatial memory for traffic forecasting}, Expert Systems with Applications 250 (2024) 123884.
\newblock \href {http://dx.doi.org/https://doi.org/10.1016/j.eswa.2024.123884} {\path{doi:https://doi.org/10.1016/j.eswa.2024.123884}}.
\newline\urlprefix\url{https://www.sciencedirect.com/science/article/pii/S0957417424007504}

\bibitem{RN28}
J.~Zhang, Y.~Zheng, D.~Qi, R.~Li, X.~Yi, Dnn-based prediction model for spatio-temporal data, in: Proceedings of the 24th ACM SIGSPATIAL international conference on advances in geographic information systems, 2016, pp. 1--4.

\bibitem{RN29}
Z.~Wu, S.~Pan, G.~Long, J.~Jiang, X.~Chang, C.~Zhang, Connecting the dots: Multivariate time series forecasting with graph neural networks, in: Proceedings of the 26th ACM SIGKDD international conference on knowledge discovery and data mining, 2020, pp. 753--763.

\bibitem{RN30}
A.~Grover, J.~Leskovec, node2vec: Scalable feature learning for networks, in: Proceedings of the 22nd ACM SIGKDD international conference on Knowledge discovery and data mining, 2016, pp. 855--864.

\bibitem{RN31}
J.~W.~C. Lint, S.~Hoogendoorn, H.~Zuvlen, Freeway travel time prediction with state-space neural networks: Modeling state-space dynamics with recurrent neural networks, Transportation Research Record 1811.
\newblock \href {http://dx.doi.org/10.3141/1811-04} {\path{doi:10.3141/1811-04}}.

\bibitem{RN32}
Z.~Lv, J.~Xu, K.~Zheng, H.~Yin, P.~Zhao, X.~Zhou, Lc-rnn: A deep learning model for traffic speed prediction, in: IJCAI, 2018, p. 27th.

\bibitem{RN33}
R.~Fu, Z.~Zhang, L.~Li, Using lstm and gru neural network methods for traffic flow prediction, in: 31st Youth Academic Annual Conference of Chinese Association of Automation (YAC), IEEE, 2016, pp. 324--328.

\bibitem{RN34}
A.~Prabowo, W.~Shao, H.~Xue, P.~Koniusz, F.~D. Salim, Because every sensor is unique, so is every pair: Handling dynamicity in traffic forecasting, in: Proceedings of the 8th ACM/IEEE Conference on Internet of Things Design and Implementation, 2023, pp. 93--104.

\bibitem{RN35}
B.~Pu, J.~Liu, Y.~Kang, J.~Chen, S.~Y. Philip, Mvstt: A multiview spatial-temporal transformer network for traffic-flow forecasting, IEEE transactions on cybernetics 54~(3) (2022) 1582--1595.

\bibitem{RN36}
C.~Zheng, X.~Fan, C.~Wang, J.~Qi, \href{https://ojs.aaai.org/index.php/AAAI/article/view/5477}{Gman: A graph multi-attention network for traffic prediction} (2020).
\newline\urlprefix\url{https://ojs.aaai.org/index.php/AAAI/article/view/5477}

\bibitem{RN37}
P.~Deng, J.~Liu, X.~Wang, X.~Jia, Y.~Zhao, M.~Wang, X.~Dai, \href{https://link.cnki.net/urlid/11.1826.tp.20230713.1714.009}{Stctn: a spatio-temporal causal representation learning method based on temporal bias adjustment and spatial causal transition}, Chinese Journal of Computers 46~(12) (2023) 2535--2550.
\newline\urlprefix\url{https://link.cnki.net/urlid/11.1826.tp.20230713.1714.009}

\bibitem{RN38}
C.~Ge, S.~Song, G.~Huang, Causal intervention for human trajectory prediction with cross attention mechanism, in: Proceedings of the AAAI Conference on Artificial Intelligence, Vol.~37, 2023, pp. 658--666.

\bibitem{RN39}
X.~Luo, W.~Yin, Z.~Li, Spatio-temporal graph neural network with hidden confounders for causal forecast, in: Special Track on AI for Socio-Ecological Welfare at ICCBR2024 (ICCBR AI Track’24), 2024.

\bibitem{RN40}
Y.~Zhao, P.~Deng, J.~Liu, X.~Jia, M.~Wang, Causal conditional hidden markov model for multimodal traffic prediction, in: Proceedings of the AAAI Conference on Artificial Intelligence, Vol.~37, 2023, pp. 4929--4936.

\bibitem{RN41}
P.~Deng, Y.~Zhao, J.~Liu, X.~Jia, M.~Wang, Spatio-temporal neural structural causal models for bike flow prediction, in: Proceedings of the AAAI conference on artificial intelligence, Vol.~37, 2023, pp. 4242--4249.

\bibitem{RN42}
J.~Liu, H.~Lin, X.~Wang, L.~Wu, S.~Garg, M.~M. Hassan, \href{https://www.sciencedirect.com/science/article/pii/S1566253524000873}{Reliable trajectory prediction in scene fusion based on spatio-temporal structure causal model}, Information Fusion 107 (2024) 102309.
\newblock \href {http://dx.doi.org/https://doi.org/10.1016/j.inffus.2024.102309} {\path{doi:https://doi.org/10.1016/j.inffus.2024.102309}}.
\newline\urlprefix\url{https://www.sciencedirect.com/science/article/pii/S1566253524000873}

\bibitem{RN43}
B.~Jing, D.~Zhou, K.~Ren, C.~Yang, Casper: Causality-aware spatiotemporal graph neural networks for spatiotemporal time series imputation, in: the 33rd ACM International Conference on Information and Knowledge Management (CIKM ’24), 2024.

\bibitem{RN44}
L.~Yao, Z.~Chu, S.~Li, Y.~Li, J.~Gao, A.~Zhang, A survey on causal inference, ACM Transactions on Knowledge Discovery from Data (TKDD) 15~(5) (2021) 1--46.

\bibitem{RN45}
J.~Zeng, G.~Zhang, C.~Rong, J.~Ding, J.~Yuan, Y.~Li, Causal learning empowered od prediction for urban planning, in: Proceedings of the 31st ACM International Conference on Information and Knowledge Management, 2022, pp. 2455--2464.

\bibitem{RN46}
S.~Deng, H.~Rangwala, Y.~Ning, Robust event forecasting with spatiotemporal confounder learning, in: Proceedings of the 28th ACM SIGKDD Conference on Knowledge Discovery and Data Mining, 2022, pp. 294--304.

\bibitem{RN47}
K.~Takeuchi, R.~Nishida, H.~Kashima, M.~Onishi, Causal effect estimation on hierarchical spatial graph data, in: Proceedings of the 29th ACM SIGKDD Conference on Knowledge Discovery and Data Mining, 2023, pp. 2145--2154.

\bibitem{RN48}
J.~Ma, Y.~Dong, Z.~Huang, D.~Mietchen, J.~Li, Assessing the causal impact of covid-19 related policies on outbreak dynamics: A case study in the us, in: Proceedings of the ACM Web Conference 2022, 2022, pp. 2678--2686.

\bibitem{RN49}
J.~Pearl, D.~Mackenzie, The Book of Why: The New Science of Cause and Effect, Basic Books, Inc., 2018.

\bibitem{RN50}
S.~He, Q.~Luo, R.~Du, L.~Zhao, G.~He, H.~Fu, H.~Li, \href{https://www.sciencedirect.com/science/article/pii/S0378437123004685}{Stgc-gnns: A gnn-based traffic prediction framework with a spatial–temporal granger causality graph}, Physica A: Statistical Mechanics and its Applications 623 (2023) 128913.
\newblock \href {http://dx.doi.org/https://doi.org/10.1016/j.physa.2023.128913} {\path{doi:https://doi.org/10.1016/j.physa.2023.128913}}.
\newline\urlprefix\url{https://www.sciencedirect.com/science/article/pii/S0378437123004685}

\end{thebibliography}







\end{document}